\documentclass{IEEEtran}
\usepackage{cite}
\usepackage{amsmath,amssymb,amsfonts}
\usepackage{algorithmic}
\usepackage[usenames,dvipsnames]{xcolor}
\usepackage{graphicx}
\usepackage{booktabs}
\usepackage{textcomp}
\usepackage{hyperref}
\usepackage{tabularx}

\usepackage{caption}
\usepackage{subcaption}
\usepackage{multirow}

\begin{document}

\title{Understanding Environmental Posts: Sentiment and Emotion Analysis of Social Media Data}

\author{
\IEEEauthorblockN{
Daniyar Amangeldi\IEEEauthorrefmark{1},
Aida Usmanova\IEEEauthorrefmark{2},
Pakizar Shamoi\IEEEauthorrefmark{3}}
\\
\IEEEauthorblockA{
\IEEEauthorrefmark{1}\IEEEauthorrefmark{3}School of Information Technology and Engineering, \\
Kazakh-British Technical University, Almaty, Kazakhstan\\
Email: \IEEEauthorrefmark{3}p.shamoi@kbtu.kz\\
}
\IEEEauthorblockA{\IEEEauthorrefmark{2}Institute of Information Systems \\
Leuphana University Lüneburg, Lüneburg, Germany\\
}
}

\maketitle
\IEEEpeerreviewmaketitle

\begin{abstract}
%

Social media is now the predominant source of information due to the availability of immediate public response. As a result, social media data has become a valuable resource for comprehending public sentiments. Studies have shown that it can amplify ideas and influence public sentiments. This study analyzes the public perception of climate change and the environment over a decade from 2014 to 2023. Using the Pointwise Mutual Information (PMI) algorithm, we identify sentiment and explore prevailing emotions expressed within environmental tweets across various social media platforms, namely Twitter, Reddit, and YouTube. Accuracy on a human-annotated dataset was 0.65, higher than Vader's score but lower than that of an expert rater (0.90). Our findings suggest that negative environmental tweets are far more common than positive or neutral ones. Climate change, air quality, emissions, plastic, and recycling are the most discussed topics on all social media platforms, highlighting its huge global concern. The most common emotions in environmental tweets are fear, trust, and anticipation, demonstrating public reactions' wide and complex nature. By identifying patterns and trends in opinions related to the environment, we hope to provide insights that can help raise awareness regarding environmental issues, inform the development of interventions, and adapt further actions to meet environmental challenges.




\end{abstract}

\begin{IEEEkeywords}
sentiment analysis, emotion analysis, social media, public perception, climate change, global warming, pointwise mutual information, Twitter, Reddit, YouTube.
\end{IEEEkeywords}
\maketitle

\section{Introduction}
\label{sec:introduction}

Environmental issues are among the most pressing challenges facing society today. In 2018, the United Nations IPCC issued a report warning of a climate change catastrophe within 12 years \cite{report}. The crucial need for environmental sustainability in the current era of climate change is stated by recent reports \cite{report} and research on sustainability \cite{sust}.

Social media platforms are crucial for environmental advocacy, as they enable individuals and organizations to share information, raise awareness, and mobilize support for common goals. Analyzing social media data involves exploring thoughts and opinions on various domains~\cite{Adwan2020}. 

Analyzing social media data is useful to understand people's opinions on various topics. Understanding the sentiment and emotion expressed in environment-related posts is important for several reasons. On social media, sentiment analysis can identify key issues and concerns and reveal patterns of public opinion and attitudes towards environmental issues. Analyzing sentiment and emotion can help identify factors that encourage participation in environmental discussions on social media. Therefore, this aims to address the gap in knowledge regarding the sentiment and emotion expressed in the comments related to the environment.


This study aims to reveal the public perception of environmental problems. The objective will be achieved by collecting data from popular social media platforms spanning over the last decade, including Twitter, Reddit, and YouTube. The textual data will be analyzed to understand the prevailing emotions related to environmental issues over the past ten years and the factors influencing public attitudes toward ecological awareness. Additionally, the study explores whether using specific social media networks impacts the emotional background of users.

We aim to answer the following questions using textual data from Twitter, Reddit, and YouTube in the period from 2013 to 2023:
\begin{itemize}
    \item How has the public perception of environmental problems changed over the decade of data?
    \item What were the prevailing emotions and topics associated with this change?
    \item What could possibly affect the attitude toward global warming and ecology problems awareness in social media? 
    \item Does the utilized social media network influence emotional background and promote distinct behavior?
\end{itemize}

The main contributions of the study may be summarised as follows:
\begin{itemize}
    \item \textit{Use of Multiple Social Media Platforms.} We provide a comprehensive analysis of user opinions and discussions on environment-related topics by utilizing data from multiple social media platforms like Reddit, YouTube, and Twitter. The multi-platform approach enables one to account for each platform's diverse user demographics and communication styles.
    \item \textit{Analysis of Posts Over a Decade. }Analyzing posts over a decade allows us to examine trends and changes in public opinion regarding the environment, ecology, and global warming. We captured the evolution of discussions, the impact of significant events or policy changes, and the shifting attitudes and awareness among the online community.
    \item \textit{Emotion Analysis besides Sentiments.} The emotion analysis complements the sentiment analysis to provide a deeper understanding of the emotional experiences and affective reactions associated with environmental discussions.
    \item \textit{Holistic Understanding of Public Perception.} Analysis was done using \textbf{popular} posts from these social networks.
\end{itemize}

The paper has been structured in the following way. This Introduction is Section I. Section II contains an overview of the literature on sentiment analysis and opinion-mining research. Section III is concerned with the methodology used for this study. Data collection and description are also covered there. Section IV presents the findings of the research. Next, the Discussion is presented in Section V. Finally, Section VI provides concluding remarks and recommendations for future enhancements to the methodology.

\section{Related Work}

\subsection{Opinion Mining in Social Media}
Social media has become a significant platform for public discussions and opinions on various topics, including ecology and the environment. It is capable of shaping public opinions and amplifying ideas. 


Sentiment Analysis, also known as Opinion mining, is one of the essential methods for understanding public views and getting insights into current trends. Such analysis has proven useful for further decision-making in various domains.
Studies like \cite{Kumar2016Opinion} and \cite{Jumadi2016Opinion} classified public opinions into negative and positive, based on Amazon reviews and Twitter comments, respectively. 
Sentiment analysis conducted by \cite{Perera2017Opinion} and \cite{Raut2014Opinion} studied restaurant reviews and hotel reviews, respectively to generate personalized review recommendations. \cite{Jeyapriya2015ExtractingOpinion} analyzed product reviews to acquire valuable information for marketing analysis. Another related work uses a similar approach for monitoring YouTube movie reviews \cite{6487473}.
\cite{Iddrisu2023SarcasticSentiments} developed a framework that supports airlines in addressing customer complaints and improving services during global events like the COVID-19 pandemic through social media sentiment analysis focusing on sarcasm detection.
\cite{Almerekhi2022Toxicity} analyzed 2 billion posts and comments from Reddit to identify toxic comments. The authors hope to bring more awareness to the online harassment problem experienced by many people nowadays and potentially prevent toxic behavior on social networks. 
The other recent study analyzed public sentiment regarding the vegan diet using Twitter data \cite{peerj}. It finds that, despite some persistent fears associated with veganism, public perception has a growing positive trend. These insights have important implications for health programs, government initiatives, and efforts to reduce veganism-related negative emotions.

Studies like \cite{Pratama2022MediaCoverage} and \cite{Faruk2022Perception} analyzed public reactions to COVID-19, providing insights into sentiment patterns during the pandemic. A paper conducted by \cite{Giachanou2016SenitimentSpikes} highlights the significance of monitoring public sentiment in social media for decision-making processes and emphasizes the importance of understanding the causes of sentiment spikes, with a focus on extracting relevant topics using the Latent Dirichlet allocation (LDA) method.

A study conducted by\cite{Shan2021} performed a sentiment analysis on Weibo posts to track people's emotional responses towards river pollution. The research demonstrates the potential of social media data for tracking emotional responses to environmental issues. The findings indicate that people tend to express more negative emotions than positive ones when discussing river pollution.

A recent study \cite{Wang2021} examined the potential of using Twitter data to detect air pollution in urban areas. The study introduced an Extended Temporary Memory (ETM) approach for air quality forecasting, which was compared to existing methods using daily data collected throughout 2019.
The results demonstrate that 4.1\% of threads on social media related to air pollution and the frequency of these terms were highly associated with air quality levels. The proposed ETM approach with sentiment analysis outperformed other prediction systems with the highest efficiency. 
This indicates that social media can function as an early warning system for natural disasters, with users providing real-time information and expressing concerns about the impact on their daily lives.

The literature on sentiment analysis of social media data related to environmental issues has shown that sentiment analysis can provide insights into the public's attitudes and emotions toward various environmental topics. \cite{Sluban2015} found that sentiment analysis aligns with attitudes towards renewable energy, sustainability, and pollution. In addition, sentiment analysis has shown negative sentiments towards topics such as $CO_2$ and fracking. These findings suggest that sentiment analysis can reasonably indicate public sentiment toward environmental topics.

Previous studies performed sentiment analysis of environmental issues on generic, domain-independent textual data. Such an approach lacks domain and context-specific training, which could limit capturing all public sentiments towards environmental issues. One of the limitations of analyzing emotions on social media is the difficulty in identifying the root causes of certain types of emotions due to the character limit of posts. Although sentiment analysis can offer valuable insights into the general emotional responses of the public toward environmental issues, it may not be sufficient to provide a comprehensive understanding of the underlying reasons behind these emotions.

Further research is needed to develop domain-specific sentiment analysis models that better capture public sentiment towards environmental issues. Improved understanding of public perception can inform environmental decision-making. Therefore, this study aims to investigate people's emotional responses to environmental issues by analyzing social media data, specifically Twitter, Reddit, and YouTube.

\subsection{Sentiment Analysis Methods}

Sentiment analysis methods fall into two categories: Lexicon-based and Machine Learning (ML)-based approaches \cite{sentimentReview1}. \cite{sentimentReview2, sentimentReview3}. Lexicon-based methods can further be split into dictionary-based techniques and corpus-based methods. These methods rely on predefined dictionaries to assess sentiment based on positive, negative, or neutral words or/and employ statistical models based on large text datasets to understand the context. ML-based methods can be divided into Supervised and Unsupervised Learning. Supervised Learning uses labeled data to train classification models like SVM, KNN, DTC, and LR. Unsupervised Learning, on the other hand, uncovers patterns in data using clustering, topic modeling, and mapping algorithms \cite{sentimentReview4, sentimentReview5, sentimentReview6}.

Even though there are other techniques to building sentiment analysis models, we picked the point-wise mutual information (PMI) approach (it is Lexicon-based) proposed by ~\cite{so,peter1,peter2,peter3} for its interpretability and robustness to statistical bias in small sample sizes \cite{peerj}.  In NLP applications, the PMI or MI evaluates the chance of two-word co-occurrence relative to the random probability, adding greater meaning to the semantic proximity of the terms. By calculating PMI, sentiment analysis algorithms can better understand the contextual and semantic relationships between words and sentiments, thus improving the accuracy of sentiment classification and providing more nuanced insights into text data \cite{peerj}. Another reason for employing the aforementioned approach is that it will be used for feature selection in this study. The principles of this approach were initially referred to as mutual information (MI)~\cite{fano1961transmission,jurafsky2014speech}. 

Our goal in employing the strategy is to evaluate how well the words that are intended to be connected with particular sentiment classes (PMI measures) can function as features while building the sentiment classifier model. The PMI-based method of sentiment analysis is employed in a lot of research ~\cite{sentiment11,vo2012sentiment,feldman2013techniques,hamdan2015sentiment, utama, bonzanini2016mastering, kanna, pmi1, pmi2}.
\section{Methods}

\begin{figure*}[!htbp]
  \centering
  \includegraphics[width=0.8\textwidth]{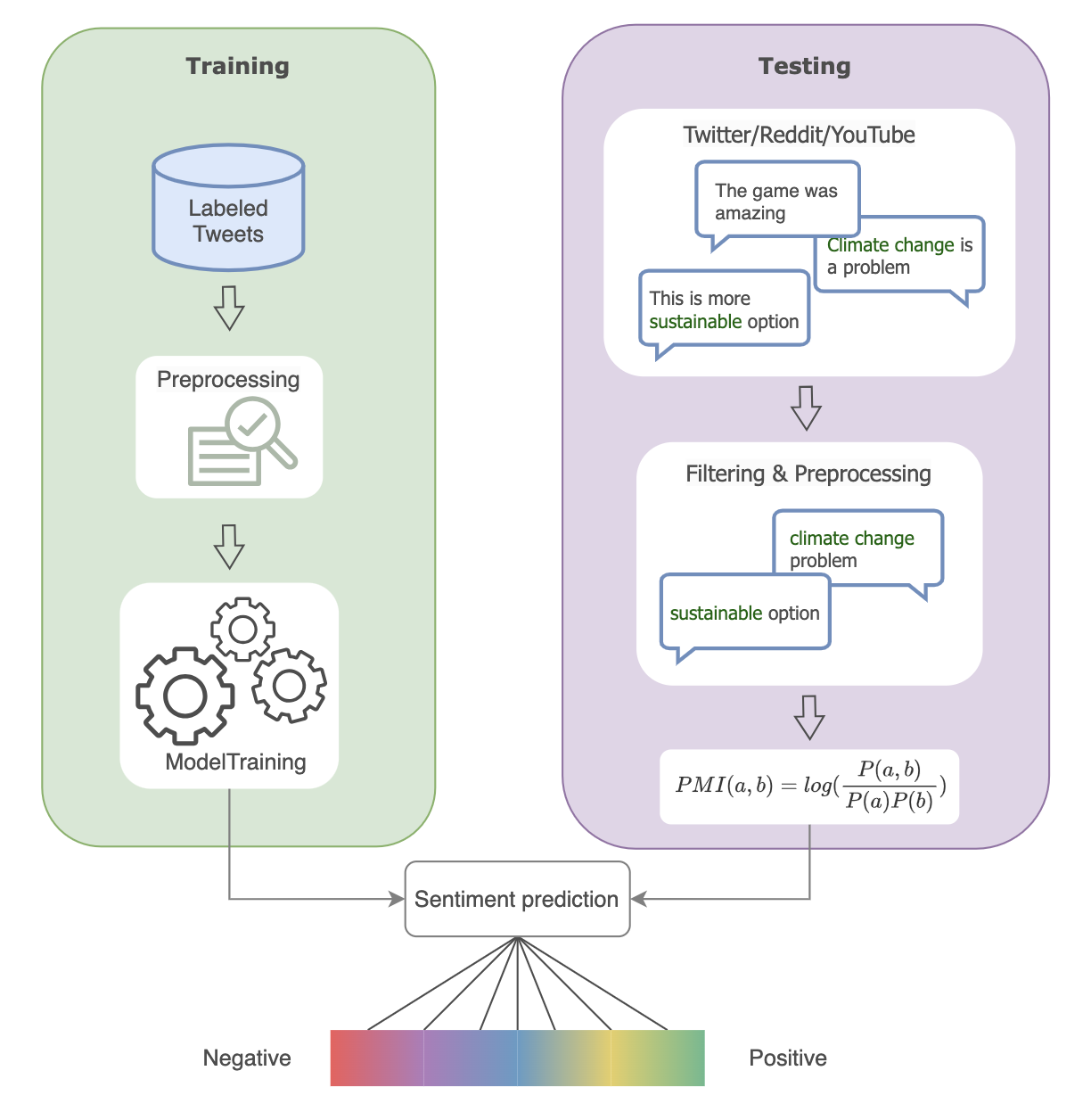}
  \caption{Study workflow: The work consists of two parts. Training is done on the labelled tweets dataset. Testing is performed on web-scraped data from Twitter, Reddit, and YouTube. The trained model is then applied to test data to generate sentiment prediction scores for each comment.}
  \label{fig:methodology}
\end{figure*}

The schematic representation of the methodology in Figure \ref{fig:methodology} shows the steps involved in the sentiment analysis process. The methodology includes steps such as data collection, pre-processing, training a model, and applying the algorithm for sentiment analysis.

\subsection{Data Collection}
To perform sentiment analysis on environmental posts, two datasets are required: the training and testing datasets. The labeled dataset contains pre-annotated tweets with sentiment labels (0 for negative and 4 for positive), which are used to train the sentiment analysis model. The unlabelled dataset is scraped from social media networks and is used for further analysis. The aim is to use the knowledge gained from the labeled training dataset to create a model that can accurately predict the sentiment of new, unlabelled tweets in the testing dataset.
\subsubsection{Traning Dataset}
\label{sec-training-data}

\begin{figure}[!htbp]
  \centering
  \includegraphics[width=0.45\textwidth]{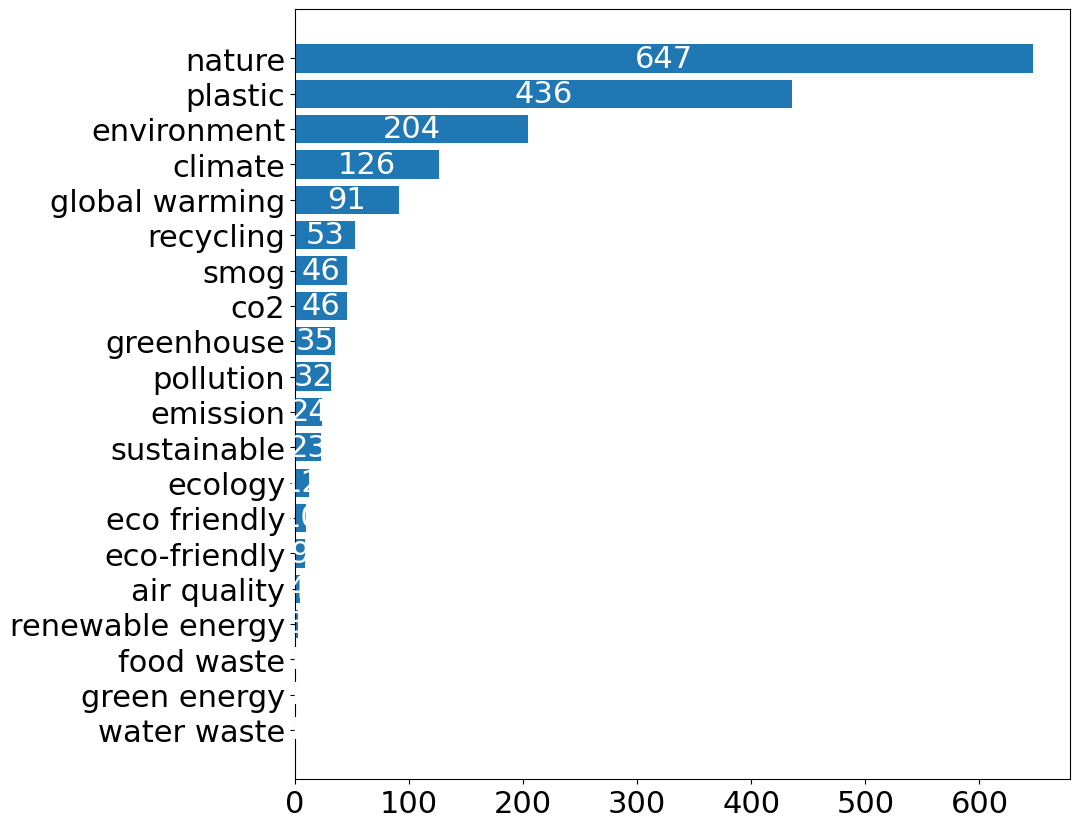}
  \caption{Environmental tweets by the keywords for training dataset}
  \label{fig:training_tweets}
\end{figure}

To train our sentiment analysis model, Sentiment140 dataset\footnote{\url{https://www.kaggle.com/datasets/kazanova/sentiment140}} was utilized. The dataset is extensively utilized in research on sentiment analysis due to its vast size and diverse content. It comprises a vast collection of textual information gathered from Twitter, covering a wide range of topics and sentiments. The dataset consists of around 1.6 million tweets, each labeled with a sentiment polarity of either positive or negative. This makes it possible to conduct supervised training of the model.

The dataset was filtered based on the following keywords: \textit{"climate", "global warming", "environment", "nature", "pollution", "plastic", "green energy", "food waste", "water waste", "greenhouse", "recycling", "air quality", "eco-friendly", "emission", "renewable energy", "sustainable", "zero waste", "carbon dioxide", "ecology", "smog", "biodiversity"}. We collected 1804 environmental tweets from a dataset of 800,000 instances, equally divided between positive and negative sentiments. The subset of filtered tweets consists of an almost equal number of tweets: 946 positive and 858 negative. Based on the Figure \ref{fig:training_tweets}, among the keywords, \textit{nature} emerges as the most prominent, dominating the conversation with 647 tweets for training, followed by \textit{"plastic"} and \textit{"environment"}.
\subsubsection{Testing Datasets}
To perform comprehensive sentiment analysis, we scraped textual information from 3 popular social media platforms, namely Twitter, Reddit and YouTube.
To ensure that the data collection process is consistent across all platforms, we used the same search keywords(mentioned in Section \ref{sec-training-data}) to gather relevant posts from each platform.


\paragraph{Twitter}
Twitter is a popular micro-blogging platform with 1.3 billion users who send out 500 million tweets daily~\cite{algren2022}. Scraping environmental tweets from the period of 2013 to 2023 was performed with snscrape\footnote{\url{https://github.com/JustAnotherArchivist/snscrape}} Python library. Tweets were filtered based on the keywords applied in filtering training data. In order to ensure a balanced representation over time and manage the size of the dataset, we limited the collection to 100 tweets per keyword each month. This approach provides a diverse and relevant set of environmental tweets that can be further analyzed and evaluated. In total, 284,440 environmental tweets for analysis and evaluation were retrieved.


\paragraph{Reddit}

To obtain a testing dataset from Reddit, the PRAW\footnote{\url{https://praw.readthedocs.io/en/stable/}} (Python Reddit API Wrapper) library was utilized to scrape environmental posts from the period 2013 to 2023. The same set of keywords was utilized for subreddit searches. To ensure temporal balance, the collection was limited to a maximum of 100 posts per keyword monthly, which resulted in 38,251 environmental Reddit posts.
\paragraph{ YouTube}
To mitigate bias and ensure more heterogeneous public feedback, we scraped comments from popular news channels on YouTube, namely Euronews, CNN, Sky News, BBC, NBC, CBC, and ABC. We collected data from YouTube by sending requests to the website and obtaining links to the relevant videos. To get the list of videos relevant to the topic, we manually selected Playlists from the channels that were related to the environment, e.g. climate change, global warming, etc. Once we got the links, we sent requests to the respective websites and scraped 100 comments containing relevant keywords in the content.
Selected videos and scraped comments were published within the 2014 and 2023 time frames. Overall, we retrieved 1998 relevant videos with 5468 relevant comments. 


\begin{figure*}[!htbp]
    \begin{subfigure}{0.33\textwidth}
        \includegraphics[width=\textwidth]{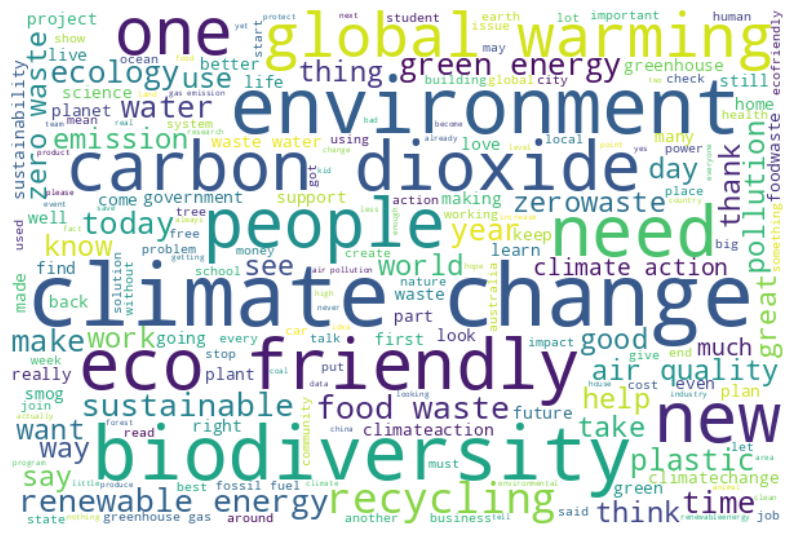}
        \caption{Twitter}
    \end{subfigure}%
    \hfill
    \begin{subfigure}{0.33\textwidth}
        \includegraphics[width=\textwidth]{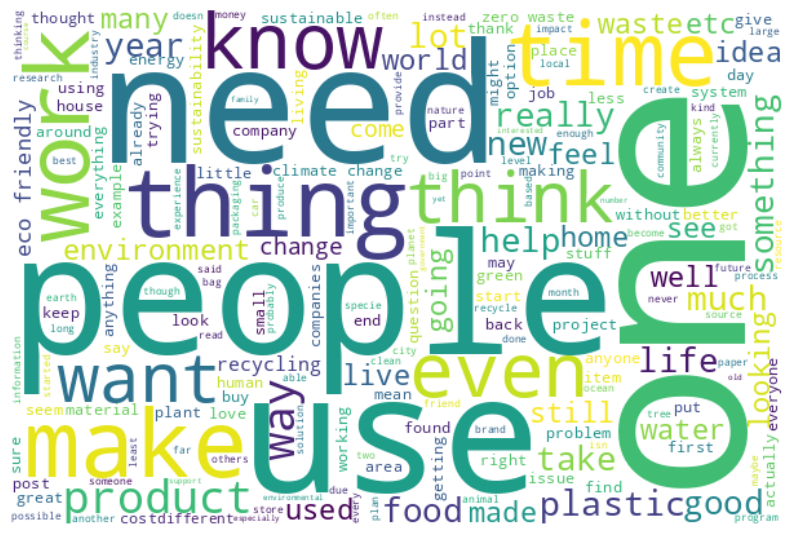}
        \caption{Reddit}
    \end{subfigure}%
    \hfill
    \begin{subfigure}{0.33\textwidth}
        \includegraphics[width=\textwidth]{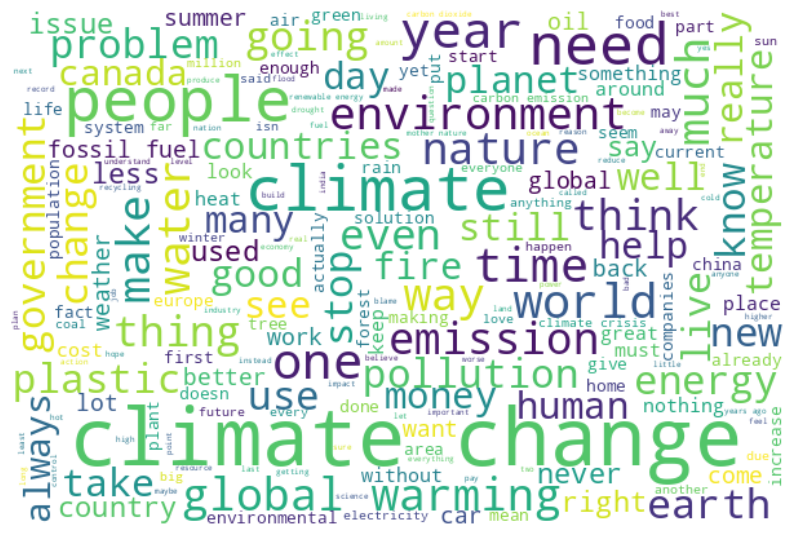}
        \caption{YouTube}
    \end{subfigure}%

    \caption{ Word clouds for \textit{popular} posts in social media.}
    \label{wordclouds}
\end{figure*}



\subsection{Description of Environmental Posts and Comments}

Table \ref{popularity} illustrates the evolution of engagement trends over the years across Twitter, Reddit, and YouTube. It becomes apparent that each platform exhibits distinct patterns in the proportion of popular posts. For instance, while Reddit consistently maintains a high percentage of popular posts (over 95\%), Twitter and YouTube steadily increased, suggesting a higher user engagement. 

Defining what constitutes a popular post across different social media platforms is a crucial aspect of our analysis because each platform has its own metrics for engagement, and setting criteria for popularity allows for consistent evaluation. Here is how we define a \textit{popular post} for:
\begin{itemize}
    \item Twitter:  at least one like
    \item Reddit: at least one upvote 
    \item YouTube: at least one like
\end{itemize}

For further analysis, we will use popular posts from our scraped data from different social network systems.

\begin{table*}[!htbp]
    \caption{Social Media Popularity}
    \centering
    \renewcommand{\arraystretch}{1.5}
     \setlength{\arrayrulewidth}{0.5mm}

    \begin{tabularx}{\textwidth}{lXXXXXXXXX}
        \hline
         & \multicolumn{3}{c}{\textbf{Twitter}} & \multicolumn{3}{c}{\textbf{Reddit}} & \multicolumn{3}{c}{\textbf{YouTube}} \\
        \cline{1-10}   
       \textbf{Year} & \textbf{Total} & \textbf{Popular}  & \textbf{Popular, \%} & \textbf{Total} & \textbf{Popular} & \textbf{Popular,\%} & \textbf{Total} & \textbf{Popular} & \textbf{Popular,\%} \\
        \hline
        2013 & 27600 & 2553 & 9.25 & 624 & 595 & 95.35 & 0.0 & 0.0 & 0.00 \\
        2014 & 27600 & 4535 & 16.43 & 705 & 671 & 95.18 & 1.0 & 1.0 & 100.00 \\
        2015 & 27600 & 5535 & 20.05 & 650 & 638 & 98.15 & 7.0 & 5.0 & 71.43 \\
        2016 & 27559 & 7648 & 27.75 & 916 & 899 & 98.14 & 40.0 & 26.0 & 65.00 \\
        2017 & 27600 & 9016 & 32.67 & 1122 & 1111 & 99.02 & 72.0 & 54.0 & 75.00 \\
        2018 & 27526 & 11019 & 40.03 & 2069 & 2030 & 98.12 & 75.0 & 38.0 & 50.67 \\
        2019 & 27075 & 11628 & 42.95 & 3785 & 3673 & 97.04 & 420.0 & 267.0 & 63.57 \\
        2020 & 26632 & 12399 & 46.56 & 4521 & 4384 & 96.97 & 205.0 & 143.0 & 69.76 \\
        2021 & 26310 & 13439 & 51.08 & 5630 & 5469 & 97.14 & 476.0 & 356.0 & 74.79 \\
        2022 & 27450 & 13159 & 47.94 & 7634 & 7425 & 97.26 & 2060.0 & 1640.0 & 79.61 \\
        2023 & 11488 & 5514 & 48.00 & 8895 & 8674 & 97.52 & 2112.0 & 1613.0 & 76.37 \\
        \hline
        \label{popularity}
    \end{tabularx}
\end{table*}

Next, Figure \ref{wordclouds} presents word clouds of text from posts scrapped from Twitter, Reddit, and YouTube. It could be noted that the absence of visually dominant words in YouTube comments within the word cloud indicates a diverse range of discussions. Conversely, in the context of environmental posts on Reddit, dominant words like \textit{"people"} in the word cloud suggest that discussions on environmental issues often intertwine with human-related aspects. This could imply a focus on how environmental problems impact people directly or indirectly, such as through policies, lifestyle changes, activism, or societal impacts. In the word cloud generated from Twitter posts about the environment, there are dominant words like \textit{"environment", "carbon dioxide", "climate change"} and \textit{"eco-friendly"}. These terms represent key focal points in discussions on Twitter regarding environmental issues.

\paragraph{Twitter}
After analyzing the language distribution of the collected tweets, we discovered that 89\% of the tweets were in English. This indicates that the English language is predominantly used in environmental discussions. Japanese accounted for approximately 3\% of the tweets, followed by French and Spanish with 2\% each. The remaining 5\% consisted of tweets in different languages, including those with fewer than 3,000 instances, among others. This language breakdown provides valuable insights into the linguistic composition of the environmental discourse captured in the testing dataset.

As per the data presented in Table \ref{tab:metrics}, it can be observed that around 60\% of the tweets in the dataset did not receive any likes, replies, retweets, or quotes. This indicates that more than half of the tweets in the dataset had limited visibility or did not resonate well with the audience, resulting in minimal engagement. However, by analyzing the tweets that received at least one like, we can gain insights into the engagement and interaction patterns of tweets that have gathered some level of attention from users.


\begin{table}[!htbp]
  \caption{Engagement Metrics of Environmental Tweets}
  \centering
   \setlength{\arrayrulewidth}{0.5mm}
    \renewcommand{\arraystretch}{1.5}
  \begin{tabularx}{\linewidth}{lrrrr}
    \hline
     & \textbf{replyCount} & \textbf{retweetCount} & \textbf{likeCount} & \textbf{quoteCount} \\
    \hline
    count & 284440.00 & 284440.00 & 284440.00 & 284440.00 \\
    mean & 0.39 & 1.41 & 4.10 & 0.10 \\
    std & 8.22 & 54.48 & 235.69 & 3.71 \\
    min & 0.00 & 0.00 & 0.00 & 0.00 \\
    50\% & 0.00 & 0.00 & 0.00 & 0.00 \\
    70\% & 0.00 & 0.00 & 1.00 & 0.00 \\
    90\% & 1.00 & 1.00 & 3.00 & 0.00 \\
    max & 2241.00 & 17840.00 & 111318.00 & 1560.00 \\
   \hline
  \end{tabularx}
  \label{tab:metrics}
\end{table}

\begin{figure}[!htbp]
  \centering
  \includegraphics[width=0.5\textwidth]{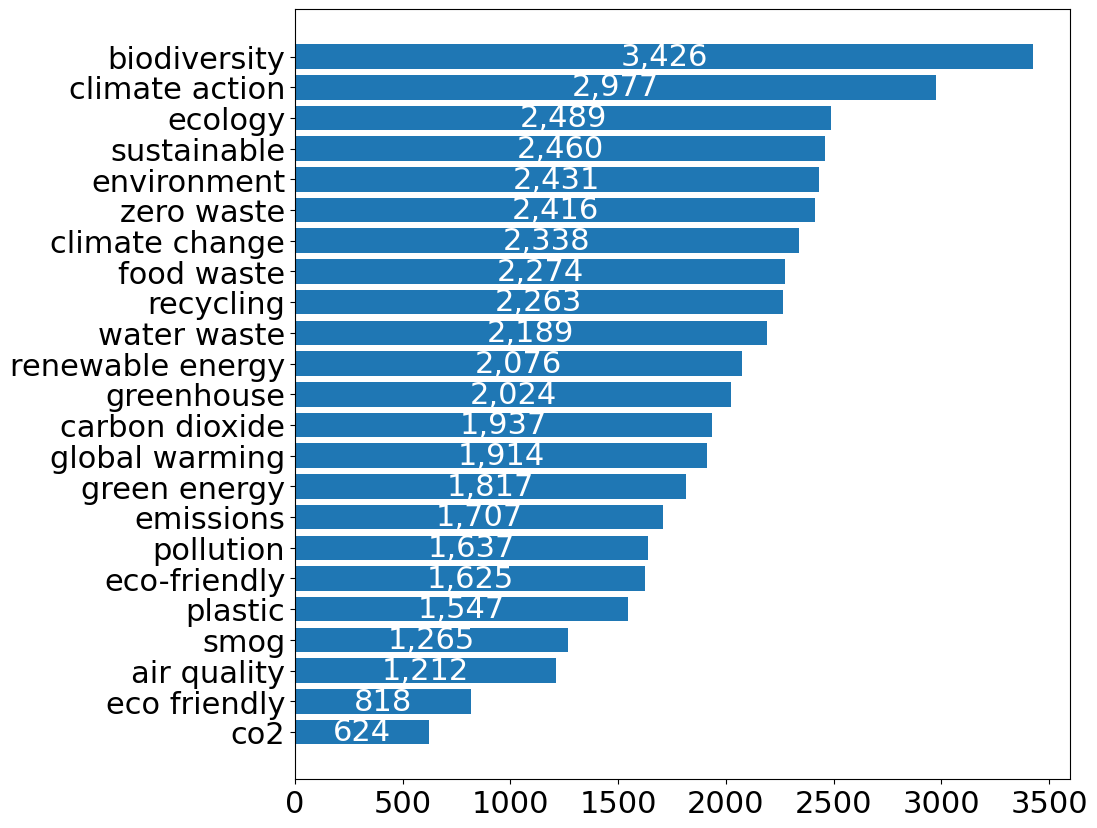}
  \caption{Popular Environmental tweets by the keywords}
  \label{fig:tweets}
\end{figure}

Figure \ref{fig:tweets} shows how often certain words were used in a collection of popular tweets about the environment. The analysis reveals that the three most commonly used words were "biodiversity", "climate action", and "ecology". These words are important because they represent key themes in discussions about the environment and sustainability. They are popular on Twitter because they align with current global environmental concerns and sustainability efforts. The fact that these words are frequently used shows that people are recognizing the need to protect biodiversity and take action to maintain ecology. This makes them highly relevant in environmental discussions.

\begin{figure}[!htbp]
  \centering
  \includegraphics[width=0.5\textwidth]{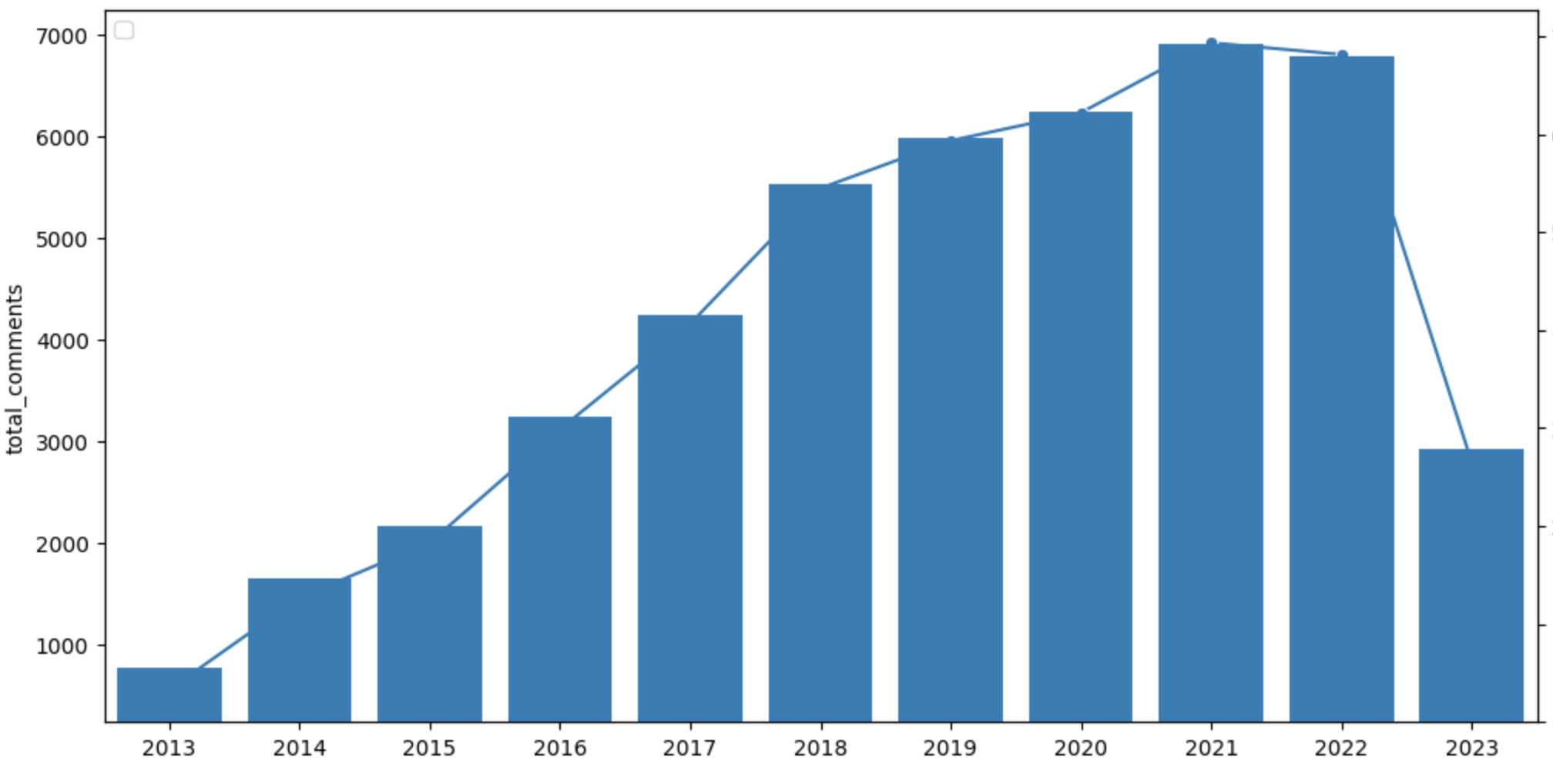}
  \caption{Number of Popular Environmental Tweets Over Time}
  \label{fig:tweets_yearly}
\end{figure}

The number of popular environmental tweets has been increasing over the years. According to Figure \ref{fig:tweets_yearly}, the count has risen from 556 tweets in 2013 to 6,916 tweets in 2021. This upward trend can be attributed to the growing accessibility and prevalence of social media platforms. As more people join these platforms and engage in online conversations, the opportunity to share and discuss environmental topics becomes more widespread. This, in turn, contributes to the overall increase in the number of environmental tweets.

\paragraph{Reddit}

\begin{figure}[!htbp]
  \centering
  \includegraphics[width=0.5\textwidth]{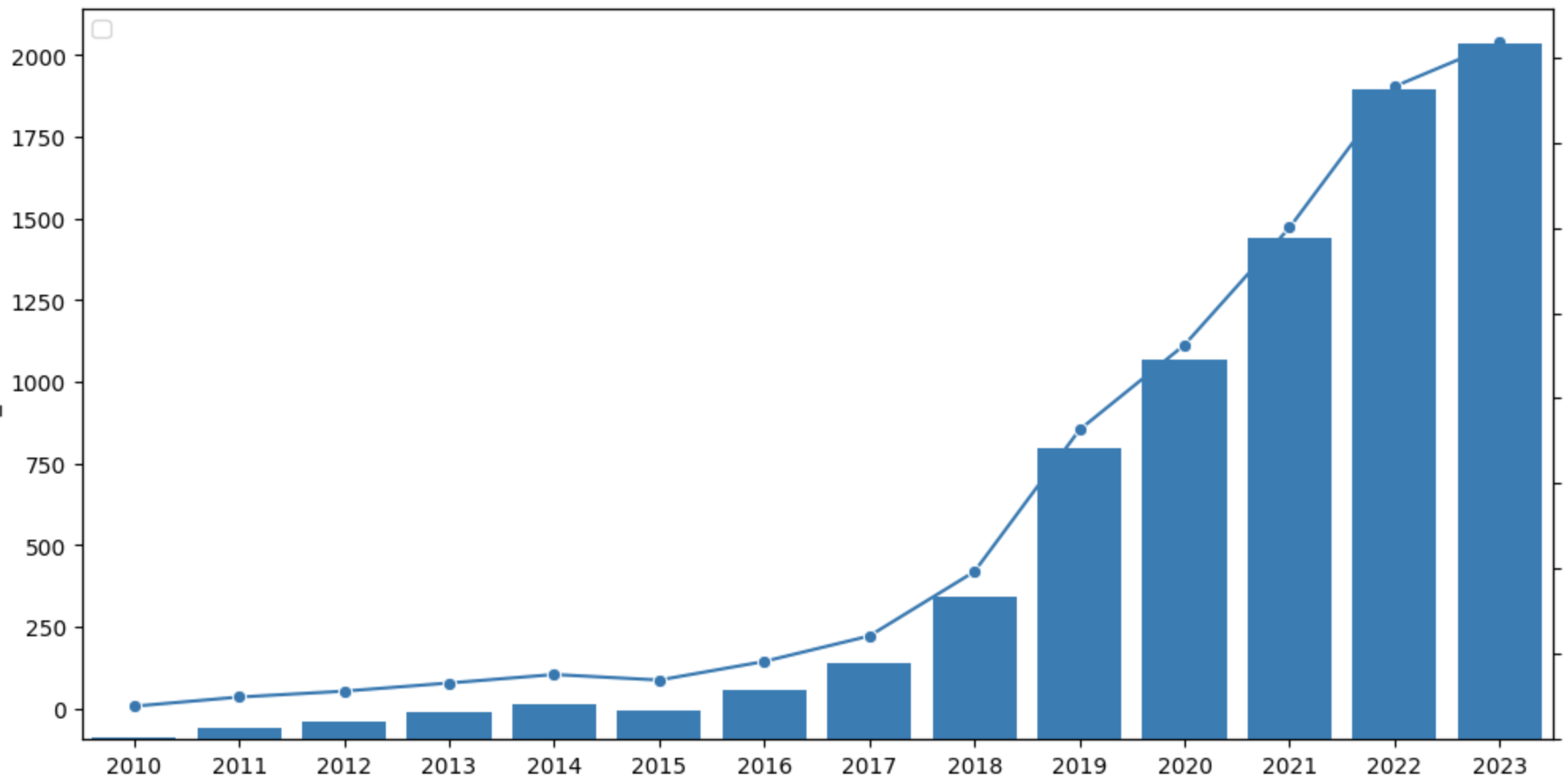}
  \caption{Number of Popular Environmental Reddit Posts Over Time}
  \label{fig:reddit_yearly}
\end{figure}

\begin{figure}[!htbp]
  \centering
  \includegraphics[width=0.5\textwidth]{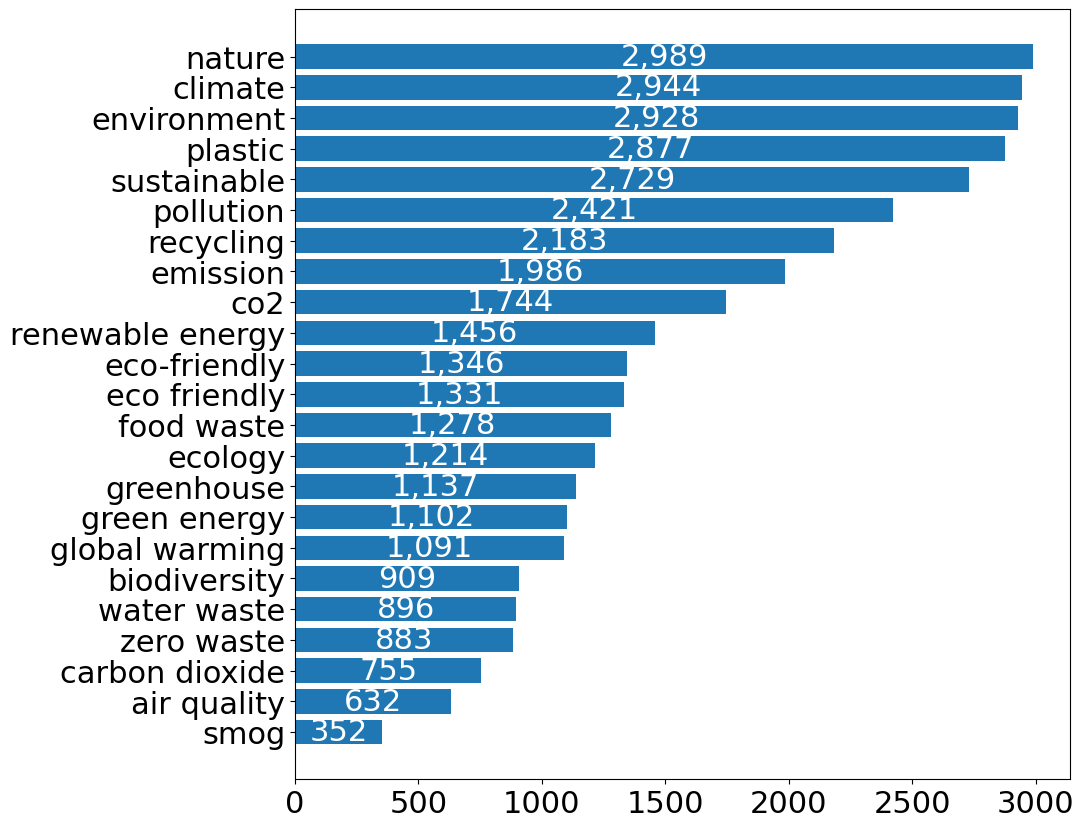}
  \caption{Popular Reddit posts by the keywords}
  \label{fig:reddit}
\end{figure}
\begin{figure*}[!htbp]
  \centering
  \includegraphics[width=0.7\textwidth]{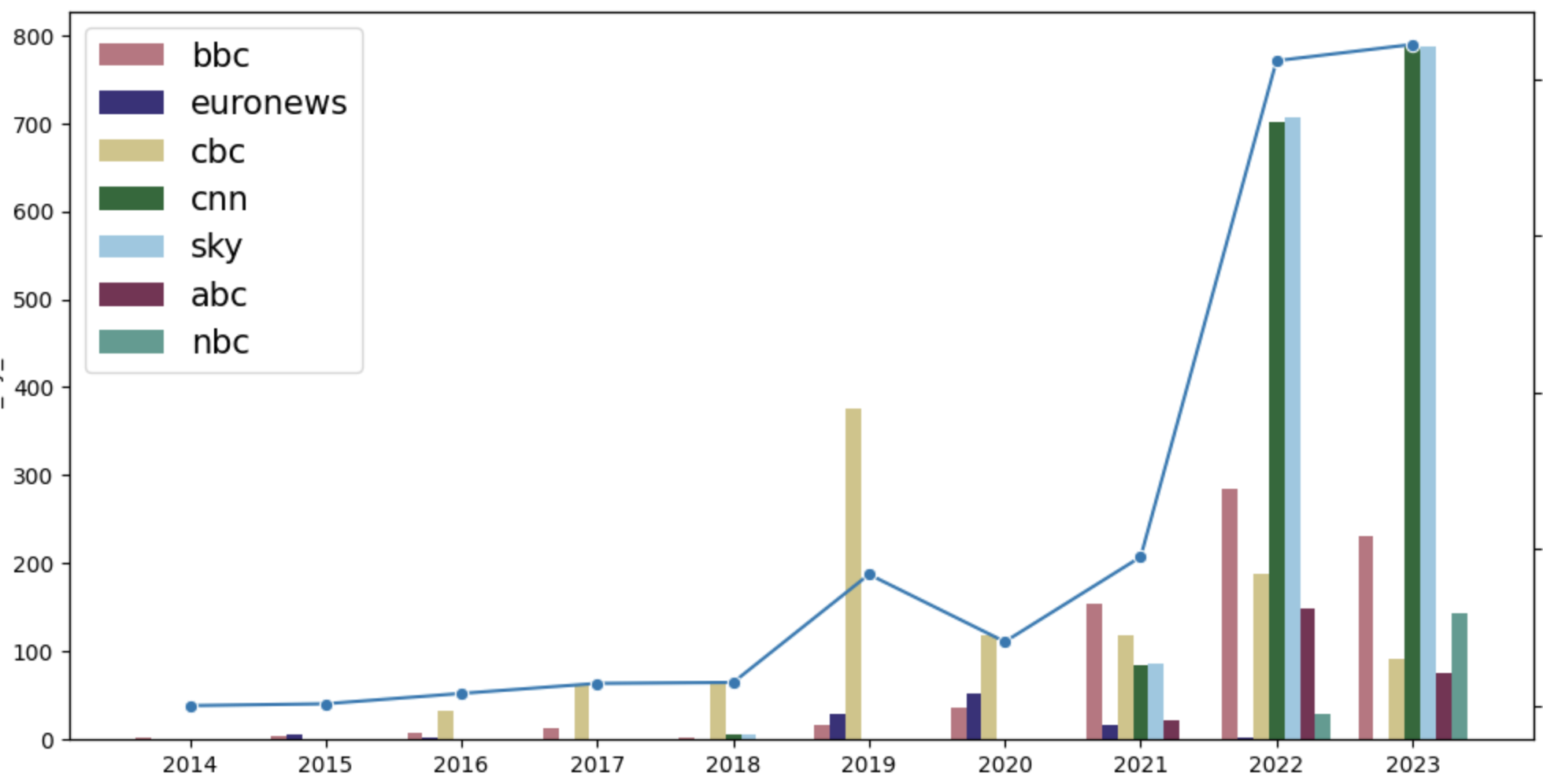}
  \caption{Number of comments on climate change and environment under YouTube videos from 2014 to 2023.}
  \label{fig:yt_comments_yearly}
\end{figure*}
According to Figure \ref{fig:reddit_yearly}, the evolution of Reddit posts over time has a noticeably upward trend, showing a significant increase in posts over the years. From a modest start in 2008 with 85 posts, the numbers gradually rose, with a more rapid increase observed after 2017. The consistent rise in the number of posts reflects a growing interest and engagement in the topic of environment. Despite Reddit scraping for only half of 2023, the popularity of posts in this period still surpasses that of previous years, indicating a sustained trend of high engagement and interaction on the platform throughout 2023.

With a total of 38,251 posts analyzed, the average number of upvotes per post stands at approximately 157. More than half of the posts received at least 15 upvotes, showing the community's interest in environmental topics. The presence of posts with exceptionally high upvotes, reaching up to 48,700, indicates the existence of standout content that captures widespread attention and engagement of society. 

Figure \ref{fig:reddit} demonstrates the frequency of popular Reddit posts by keywords. The high frequency of keywords like \textit{"climate", "environment"} and \textit{"nature"} suggests a significant Reddit focus on broad environmental topics. Additionally, terms like \textit{"sustainable"} and \textit{"renewable energy"} indicate a growing interest in ecologically clean practices within the community.

\paragraph{YouTube}




The scraped comments from YouTube were not distributed evenly.
Most comments were gathered from Sky News, CNN, and CBC channels, with 1584, 1577, and 1045 comments, respectively. The remaining four channels correspond to BBC 745, ABC 244, NBC 172, and Euronews 101 comments 
Out of 5468 gathered comments, 4151 received at least one vote, which was considered in the further analysis.

As seen from Figure \ref{fig:yt_comments_yearly} the number of discussions on environmental topics has constantly risen since 2014. There is a sudden increase in the number of comments from the CBC channel, which was likely triggered by Australian wildfires in 2019. The overall trend was then followed by an observable decline in 2020, which could be associated with the public focus shifting towards the COVID-19 pandemic. In the past two years, the numbers have drastically increased, with the environment being one of the hot topics of discussion.

\subsection{Data Analysis}
To better understand the nature of our data and remove any non-relevant information, we perform the following data analysis. The pre-processing step includes noise reduction, standardization, stop-words removal, etc. Once we get the clean data, we perform Sentiment and Emotion Analysis together with Topic Modeling.

\subsubsection{Pre-processing}
\begin{figure*}[!htbp]
  \centering
  \includegraphics[width=\textwidth]{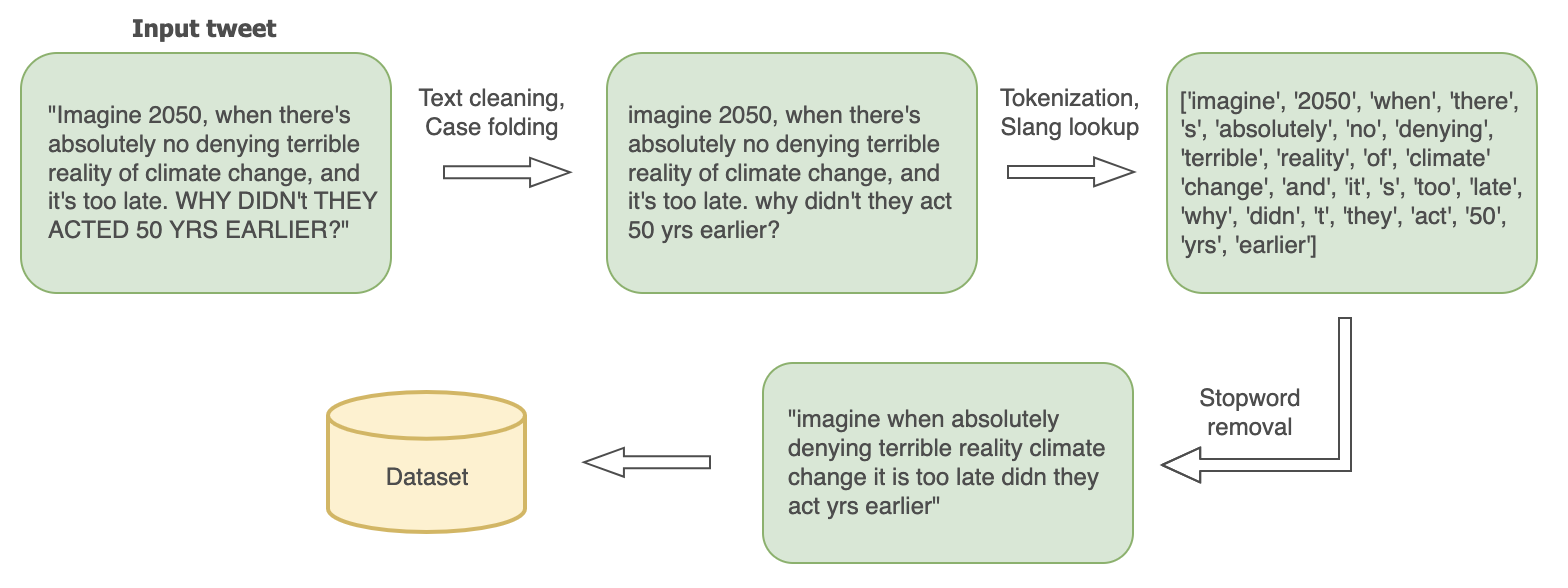}
  \caption{Comment pre-processing steps. Firstly, the texts are brought to lowercase, and various special characters are removed. This is followed by tokenization and transforming slang into full forms. As a last step, stopwords are removed from comments, creating a clean dataset.}
  \label{fig:preprocessing}
\end{figure*}

Data pre-processing involves several steps to transform raw textual data into a format suitable for analysis (see Figure \ref{fig:preprocessing}):

\begin{itemize}
   \item \textit{Cleaning}: When cleaning tweets, irrelevant elements such as URLs, special characters, hashtags, and mentions are removed to ensure only relevant content remains.

   \item \textit{Case Folding}: Text converted to lowercase to standardize text and avoid word duplication.

   \item \textit{Tokenization}: Breaking down sentences into individual words or tokens facilitates further analysis and processing by making each word a separate entity. This step also helps in removing punctuation and splitting hashtags or compound words into meaningful units.

   \item \textit{Slang Lookup}: Social media texts often contain slang words and abbreviations. To make the text easier to understand, these slang words and abbreviations are replaced with their corresponding full forms or standard equivalents. This step helps to improve the readability and comprehensibility of the text.

   \item \textit{Stopwords Removal}: Stopwords are common words in a language that don't carry significant meaning. They are removed in textual data pre-processing to reduce noise and focus on meaningful content.
   
\end{itemize}

After completing the steps of textual data pre-processing, the raw comments and posts are transformed into a clean and standardized format that is ready for further analysis and interpretation.\cite{pandey_tweetpreprocessing}.

\subsubsection{Sentiment Analysis}

In this study, we used Pointwise Mutual Information (PMI) to measure the association between words and their sentiment orientations. PMI calculates the statistical dependence between two words by comparing their co-occurrence in a given corpus with their individual occurrences. Specifically, PMI measures the logarithm of the ratio between the observed co-occurrence probability of two words and the expected probability if they were independent. This helps us understand how closely related two words are in terms of their sentiment orientation \cite{peerj} (as seen from Equation \ref{eq:pmi}).

\begin{equation}
\text{{PMI}}(w_1, w_2) = \log \left( \frac{{\text{{P}}(w_1, w_2)}}{{\text{{P}}(w_1) \cdot \text{{P}}(w_2)}} \right)
\label{eq:pmi}
\end{equation}


where $\textit{PMI}(w_1, w_2)$ represents the observed co-occurrence probability of $\textit{word}_1$ and $\textit{word}_2$, and $\textit{P}(w_1)$ and $\textit{P}(w_2)$ represent their individual occurrence probabilities.

TTo determine the semantic orientation of a word, we use the PMI scores between the target word and a set of positive ($\textit{p}$) and negative ($\textit{n}$) sentiment words. The semantic orientation ($\textit{SO}$) is calculated by subtracting the accumulated PMI scores with negative sentiment words from the accumulated PMI scores with positive sentiment words, and then dividing the result by the frequency of the target word. This formula is shown in Equation \ref{eq:so}.
\begin{equation}
\text{SO}(w) = \frac{\sum_{p \in P} \text{PMI}(w, p) - \sum_{n \in N} \text{PMI}(w, n)}{\text{word freq.}(w)}
\label{eq:so}
\end{equation}

where $\textit{P}$ and $\textit{N}$ represent positive and negative sentiment words, respectively, and $\text{word freq.}(\text{word})$ represents the frequency of the target word in the dataset.

This approach allows us to capture the sentiment associations of individual words based on their co-occurrence patterns with positive and negative sentiment words, providing insights into the semantic orientation of the words in our sentiment analysis.

Finally, the sum of individual sentiment scores results in a sentiment score of a comment (as seen from Equation \ref{eq:sentiment}), which is a numerical measure, where a positive score indicates a positive sentiment, a negative score suggests a negative sentiment and a score of zero shows a lack of strong emotional tone.

\begin{equation}
    \text{CommentSentiment(C)} = \sum_{{c \in C}} \text{SO(c)}
    \label{eq:sentiment}
\end{equation}


\subsubsection{Emotion Analysis} 

While sentiment analysis focuses on the polarity of opinions (positive, negative, neutral), emotion analysis helps to dive into the specific emotional states expressed in the posts (e.g., happiness, sadness, anger, fear, surprise). Analyzing emotional experiences and affective reactions associated with environmental discussions, enables us to capture a more nuanced understanding of public opinion. NRCLex \cite{mohammad2013nrc} is utilized to identify the emotional effect of comments. 

In this study, we decided to focus on comments that were classified as negative in the Sentiment Analysis step. NRCLex is applied to observe the emotion distribution among filtered comments. NRCLex contains 11 emotions, out of which we used only 8, removing \textit{positive}, \textit{negative}, and \textit{anticip}. The emotion intensity range is between 0 and 1. Each comment consists of a combination of various emotions with one prevailing emotion. We considered an emotion prevailing if the intensity score was over 0.25.

We use the following emotions from NRCLex: \textit{fear, anger, anticipation, trust, surprise, sadness, disgust, joy}. We excluded positive and negative emotions as we used our sentiment classifier for this purpose.

\subsubsection{Topic Modeling}
Topic Modeling clusters information into bigger groups and helps to identify present topics in textual datasets. BERTopic \cite{grootendorst2022bertopic} is a common tool to perform Topic Modeling. The tool helps to divide textual information into meaningful clusters based on their semantic meaning. 

In our study, we applied BERTopic on all comments from each platform and comments labeled as prevailing emotions from the previous step. We separately performed Topic Modeling on comments related to \textit{fear}, \textit{trust}, and \textit{anticipation} emotions.

\section{Dataset Annotation}
\begin{figure*}[!htbp]
\centering
\includegraphics[width=\textwidth]{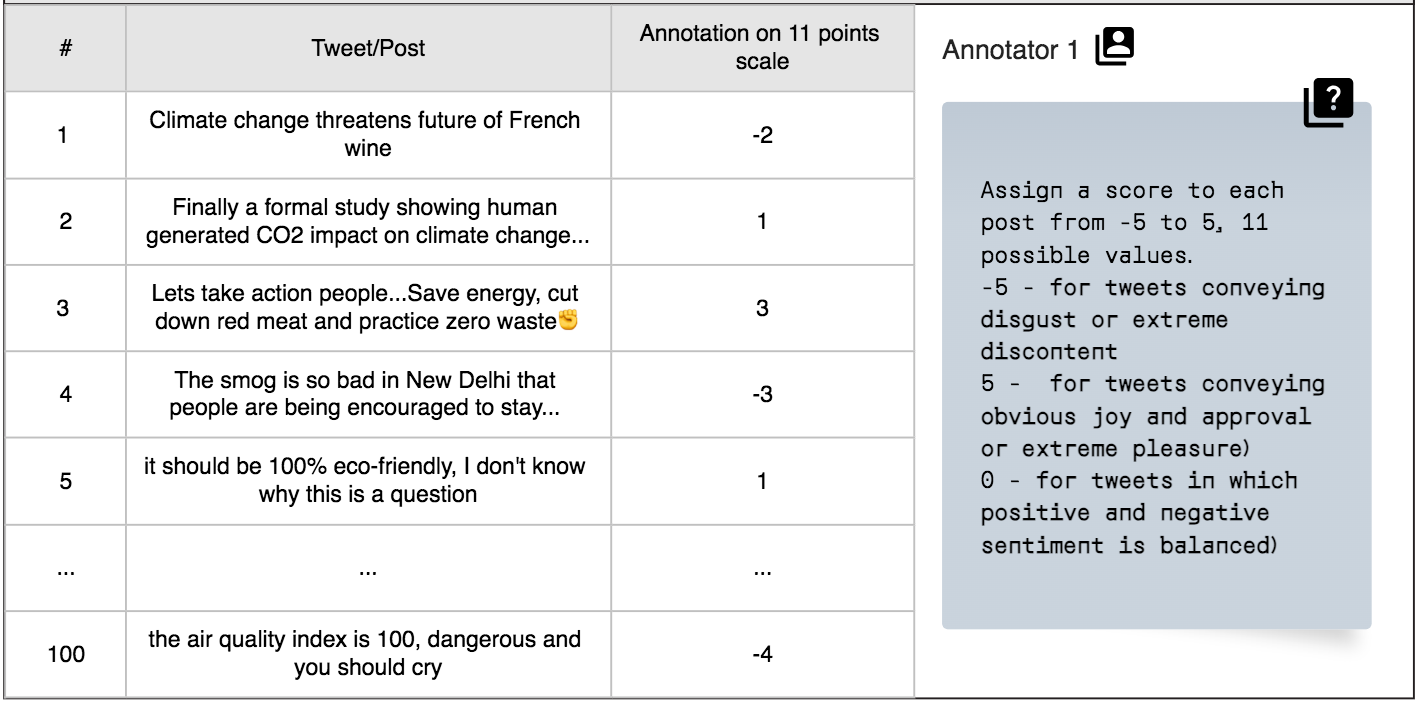}
\caption{An example form (Google Sheets) featuring comments or posts from Twitter/Reddit/Youtube that was given to each annotator individually for annotation. The following study's score descriptions and data annotation techniques were taken from  \cite{rosso}. The page design was taken from \cite{peerj}. The form contains 100 random posts from our unlabeled environment-related dataset. }
\label{annform}
\end{figure*}
\begin{table*}[!htbp]
    \caption{Sentiment Analysis for selected tweets based on manual annotation. Expert annotators are marked with asterisks, and they were given twice the weight.}
    \centering
    \setlength{\arrayrulewidth}{0.5mm}
    \setlength{\tabcolsep}{10pt}
    \renewcommand{\arraystretch}{1.5}
    \begin{tabularx}{\textwidth}{cccccccccccc}
    \hline
    \textbf{\#} & \textbf{Tweet} & \textbf{EA1*} & \textbf{EA2*} & \textbf{A3} & \textbf{A4} & \textbf{A5} & \textbf{A6} & \textbf{Weighted AVG} & \textbf{Annotation} & \textbf{Sentiment} \\ [1ex]
    \hline
        1 &Nestlé helps farm...& 3 & 2 & 3 & 3 & 1 & 4 & 2.63 & 1 & Positive  \\
        2 &Geology research...& -2 & 0 & 0 & -2 & -1 & -2 & -1.13 & -1 & Negative  \\
        3 &New paper demo...& 0 & 0 & 1 & -1 & 0 & 2 & 0.25 & 1 & Positive  \\
        4 &If climate change...& 4 & 5 & 3 & 2 & 3 & 3 & 3.63 & 1 & Positive  \\
        5 &Climate change th...&  -4 & -1 & -2 & 3 & -2 & 3 & -1 & -1 & Negative  \\
       ... & ... & ... & ... & ...  & ...  & ...  & ...  & ...  & ...   \\
       100 & Supporting the tra...& 5 & 4 & 5 & 4 & 3 & 4 & 4.25 & 1 & Positive  \\

    \hline
    \end{tabularx}
    \label{tab:ann}
\end{table*}

\begin{figure}[!htbp]
\centerline{\includegraphics[width=0.5\textwidth]{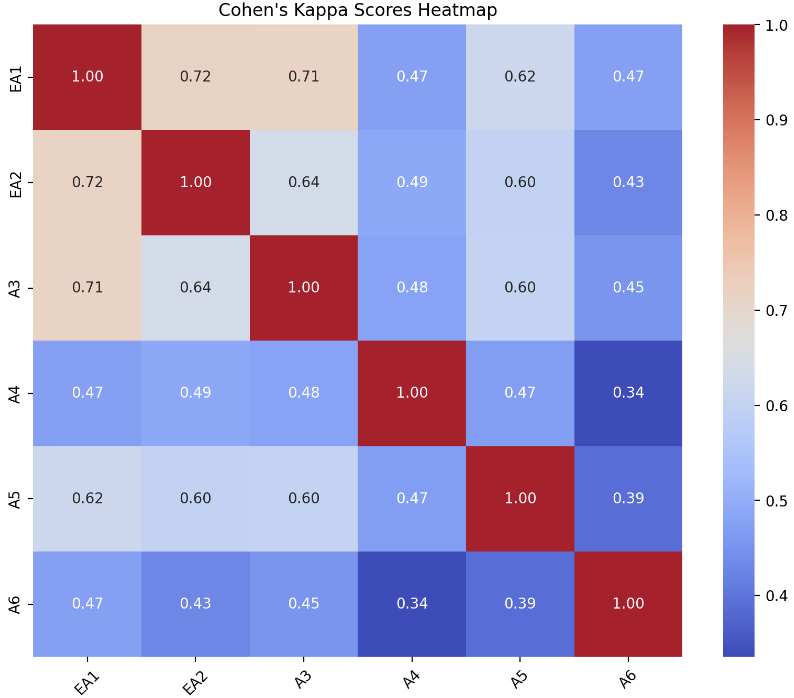}}
\caption{The heatmap showing the correlation between the ratings of different annotators - two expert annotators and four general annotators. Cohen's Kappa for each pair of annotators.}
\label{heatmap}
\end{figure}


We annotated the dataset because there were no human-labeled or classifier-trained tweets in this environment context. We randomly sampled 100 tweets from our previously mentioned dataset, focusing on environment-related content.

A sentiment analysis dataset annotated on an 11-point scale implies that each instance in the dataset is assigned a sentiment label on a scale ranging from -5 to 5. 

Six human subjects performed the annotation process. Among them, there was one expert in ecology and one PhD in ecology, so they were given a double weight since they better understood the context and sentiment conveyed by specific terms within discussions about climate change. 

All six participants in our annotation process have formally passed through the informed consent procedure, demonstrating their understanding and willingness to participate in the study. Figure~\ref{annform} shows a screenshot of the Google Sheets form. The 11-point annotation approach proposed in \cite{rosso} was used to annotate the test dataset. Each annotator was needed to assign a score to a tweet's opinion based on a perceived value ranging from -5 (showing significant discontent) to +5 (for exceptionally positive tweets). 

The total sentiment score for each tweet was derived as a weighted average of all six annotators (A), with experts (Expert Annotators, EA)receiving a twofold weighting, as mentioned above.

\begin{equation}
    \text{{Total Sentiment Score}}_{\text{{tweet}}} = \frac{{\sum_{i=1}^{n} w_i \times \text{{Annotator}}_i}}{{\sum_{i=1}^{n} w_i}}
\end{equation}

where \(w_i\) is the weight assigned to annotator \(i\), and \(\text{{Annotator}}_i\) represents the sentiment score assigned by annotator \(i\),\(n\) is the number of annotators.

For expert annotators (EA) receiving a twofold weighting, non-expert annotators (A) receiving weight 1:
\[
w_{\text{{EA}}} = 2, w_{\text{{A}}} = 1
\]

We applied the strategies described in \cite{rosso}: if 60\%  or more of annotator labels are considered outliers, the annotator judgments are removed from the job. We utilize the formula \equationautorefname~\ref{eq:score}  to determine whether a judgment $A_i,j$ is an outlier \cite{rosso}:

\begin{equation}
\label{eq:score}
    |A_{i,j}-avg(A_{i',j})| > std_{t}(t_{j}),
\end{equation}
where $std_{t}(t_{j})$ is the standard deviation of all scores given for a tweet $t_{j}$.

As a result, no outliers were revealed since we got the following proportion of outlier labels: $EA_1 = 8\%$ , $EA_2 = 25\%$ ,  $A_4 = 29\%$ , $A_5 = 35\%$, $A_6 = 36\%$.

Each tweet in a trial dataset eventually received a positive, neutral, or negative score based on weighted average scoring (see Table~\ref{tab:ann}). With the classification threshold set at –0.1 and +0.1, the annotation process produced the following sentiment distribution: positive (44\%), neutral (0\%), and negative (56\%).


Inter-Annotator Agreement (IAA) \cite{iaaa} measures the level of agreement between multiple annotators in their assessments of environmental tweets. Specifically, we use Cohen's Kappa ($\kappa$) as a measure to quantify the level of agreement among the annotators. 

The formula for Cohen's Kappa is given by:
\begin{equation}
\kappa = \frac{P_o - P_e}{1 - P_e} \label{eq:cohen_kappa}
\end{equation}
where:
\begin{align*}
P_o &= \frac{\text{Number of agreements}}{\text{Total number of annotations}} \\
P_e &= \sum_{i} \left(\frac{\text{Total annotations by annotator } i}{\text{Total number of annotations}}\right)^2
\end{align*}

So, $P_o$ represents the observed agreement - the proportion of times the annotators agree, while $P_e$ denotes the expected agreement, the hypothetical probability of chance agreement.

First, we converted the annotations of each subject from an 11-point scale to positive (1), neutral (0), or negative(-1) to calculate agreement. Then, we calculate Pairwise Cohen's Kappa. Finally, we calculate the average Cohen's Kappa across all pairs of annotators to get an overall measure of the agreement.

The heatmap in Figure \ref{heatmap} visualizes the Cohen's Kappa scores for each pair of annotators, providing insight into their level of agreement. The values range from -1 to 1, where 1 indicates perfect agreement, 0 indicates no agreement, and -1 indicates perfect disagreement. 

The value of $\kappa $ can range from -1 (complete disagreement) to 1 (complete agreement). A value of 0 indicates that the agreement is no better than chance. The average Cohen's Kappa score across all pairs of annotators is 0.525 in our case, which suggests a \textit{moderate} level of agreement among the annotators. Experts agreement is very high.

\section{Experimental Results}
\begin{figure}[!htbp]
  \centering
  \includegraphics[width=0.8\linewidth]{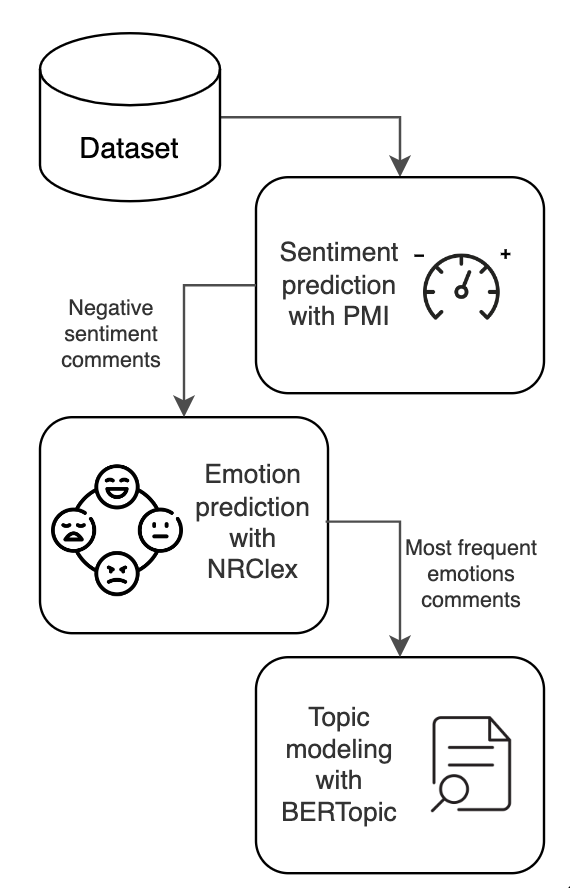}
  \caption{Analysis flow}
  \label{fig:analysis_flow}
\end{figure}
The experiments involved three stages (as seen in Figure \ref{fig:analysis_flow}): Sentiment prediction, Emotion prediction, and Topic modeling. The following sections describe each stage of the analysis process. 


\subsection{Sentiment Detection Results}
\begin{figure*}[!htbp]
    \begin{subfigure}{0.33\textwidth}
        \includegraphics[width=\textwidth]{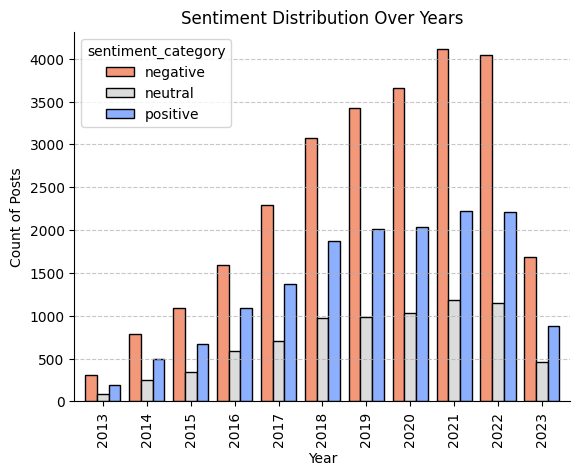}
        \caption{Twitter}
    \end{subfigure}%
    \hfill
    \begin{subfigure}{0.33\textwidth}
        \includegraphics[width=\textwidth]{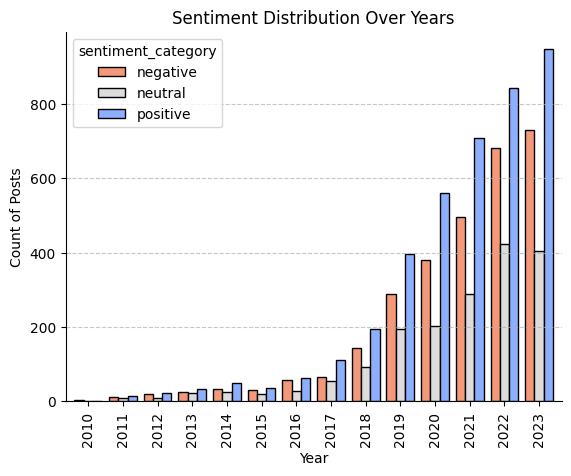}
        \caption{Reddit}
    \end{subfigure}%
    \hfill
    \begin{subfigure}{0.33\textwidth}
        \includegraphics[width=\textwidth]{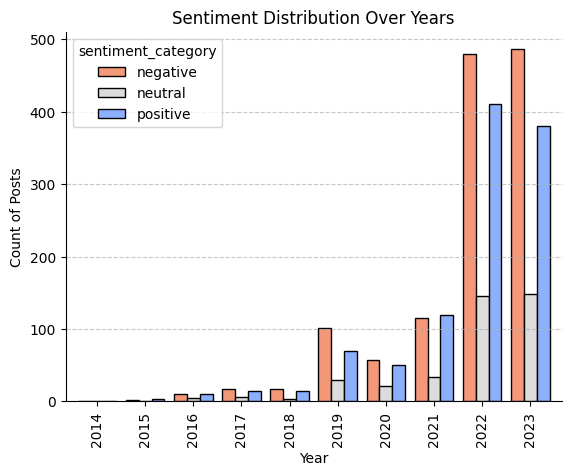}
        \caption{YouTube}
    \end{subfigure}%

    \caption{Sentiment scores distribution over the years.}
    \label{fig:sentiment-scores}
\end{figure*}
\begin{figure}[!htbp]
\centerline{\includegraphics[width=0.5\textwidth]{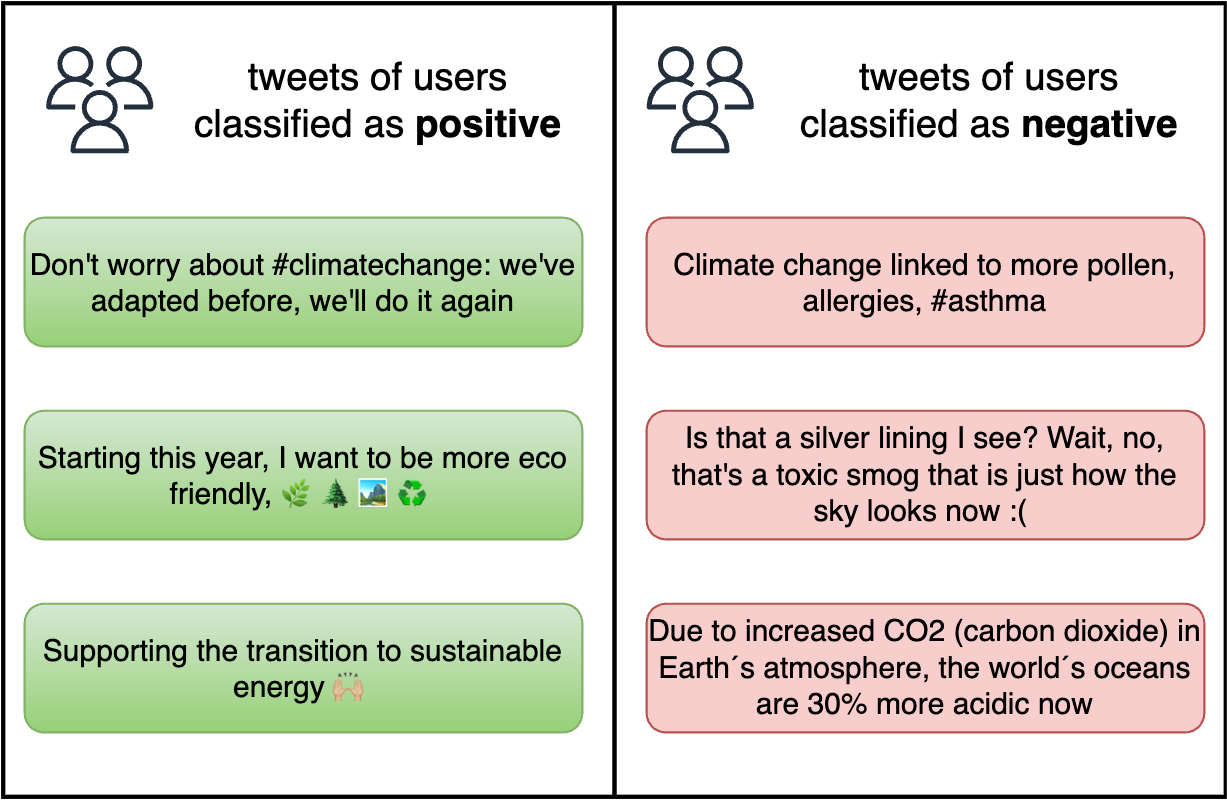}}
\caption{Examples of tweets classified as positive and negative.}
\label{tw_examples}
\end{figure}

\begin{table*}[!htbp]
    \caption{Sentiment Analysis statistical information per year.}
    \centering
    \setlength{\arrayrulewidth}{0.5mm}
    \setlength{\tabcolsep}{10pt}
    \renewcommand{\arraystretch}{1.5}
    \begin{tabularx}{\textwidth}{cccccccccc}
    \hline
    \textbf{Platform} & \textbf{Year} & \textbf{Popular} & \textbf{Positive} & \textbf{Positive, \%} & \textbf{Negative} & \textbf{Negative, \%} & \textbf{Neutral} & \textbf{Neutral, \%} & \textbf{Emotion}  \\ 
    \hline
    \multirow{11}{*}{Twitter} & 
      2014 &1530& 494 & 32.29	&783&51.18&253&16.54 & Anticipation \\
    & 2015 &2101& 671 & 31.94	&1086&51.69&344&16.37 & Surprise \\
    & 2016 &3279& 1096 & 33.42&1598&48.73&585&17.84 & Anticipation \\
    & 2017 &4364& 1366 & 31.30&2296&52.61&702&16.09 & Surprise \\
    & 2018 &5930& 1876 & 31.64&3079&51.92&975&16.44 & Fear \\
    & 2019 &6419& 2008 & 31.28&3430&53.44&981&15.28 & Fear \\
    & 2020 &6725& 2039 & 30.32&3655&54.35&1031&15.33 & Anticipation \\
    & 2021 &7513& 2226 & 29.63&4108&54.68&1179&15.69 & Surprise \\
    & 2022 &7405& 2211 & 29.86&4045&54.63&1149&15.52 & Anticipation \\
    & 2023 &3037& 886 & 29.177&1687&55.55&464&15.28 & Fear \\
    &\textbf{ Total} &\textbf{48887}& \textbf{15060} & \textbf{30.81\% }&\textbf{26079}& \textbf{53.35\% }&\textbf{7748}& \textbf{15.85\% }& \textbf{Anticipation} \\
    \midrule
    \multirow{11}{*}{Reddit} & 
      2014 & 108	& 49 & 45.37 & 34 & 31.48 & 25&  23.15& Trust \\
    & 2015	& 88 & 35 & 39.77 & 32 & 36.36 & 21& 23.86 & Fear \\
    & 2016	& 150 & 63 & 42.00 & 58 & 38.67 & 29& 19.33& Anticipation \\
    & 2017	& 230 & 111 & 48.26 & 65 & 28.26 & 54&  23.48& Fear \\
    & 2018	& 432 & 195 & 45.14 & 144 & 33.33 & 93 & 21.53& Anger \\
    & 2019	& 881 & 395 & 44.84 & 290 & 32.92 & 196& 22.25& Fear \\
    & 2020	& 1144 & 559 & 48.86 & 381 & 33.30 & 204& 17.83& Surprise \\
    & 2021	& 1494 & 708 & 47.39 & 496 & 33.20 & 290 & 19.41& Surprise \\
    & 2022	& 1947 & 843 & 43.30 & 682 & 35.03 & 422 & 21.67& Surprise \\
    & 2023	& 2079 & 946 & 45.50 & 729 & 35.06 & 404 & 19.43 & Fear \\
    &\textbf{ Total} &\textbf{8732}& \textbf{3977} & \textbf{45.55\% }&\textbf{2972}& \textbf{34.04\%} &\textbf{1783}& \textbf{20.42\%} & \textbf{Fear}\\
    \midrule
    \multirow{11}{*}{YouTube} & 
      2014& 1& 0& 0& 1& 100.00& 0& 0 & - \\
    & 2015& 5& 3& 60.00& 2& 40.00& 0& 0 & Fear \\
    & 2016& 26& 11& 42.31& 10& 38.46& 5 & 19.23 & Anticipation \\
    & 2017& 38& 15& 39.47& 17& 44.74& 6 & 15.79 & Trust \\
    & 2018& 36& 15& 41.67& 17& 47.22 & 4 & 11.11 & Fear \\
    & 2019& 201& 70& 34.83& 102& 50.75 & 29 & 14.43 & Anger \\
    & 2020& 129& 50& 38.76& 57& 44.19 & 22 & 17.05 & Trust \\
    & 2021& 269& 120& 44.61& 115& 42.75 & 34 & 12.64 & Fear \\
    & 2022& 1036& 411& 39.67& 480& 46.33 & 145& 14.00& Fear \\
    & 2023& 1015& 281& 37.54& 486& 47.88 & 148 & 14.58 & Fear \\
    &\textbf{ Total} &\textbf{2756}& \textbf{1076} & \textbf{39.04\% }&\textbf{1287}& \textbf{46.70\%} &\textbf{393}& \textbf{14.26\%} & \textbf{Fear} \\
    \hline
    \end{tabularx}
    \label{tab:main}
\end{table*}

Figure \ref{fig:sentiment-scores} and Table \ref{tab:main} demonstrate the sentiment distribution over the years on three different social network systems: Twitter, Reddit, and YouTube. The comparison indicates a distinct sentiment pattern among these platforms. Contrary to the prevalent negative sentiments found consistently on Twitter throughout the years, Reddit exhibits a rising trend in positive expressions. In contrast, YouTube portrays a slightly higher frequency of negative sentiments, indicating a different sentiment landscape within the platform. Twitter and YouTube predominantly showcase negative sentiments, suggesting a prevalence of critical or adverse expressions among their user bases. However, the noticeable increase in positive sentiments on Reddit reveals a contrasting sentiment trend, showcasing an evolving and comparatively more optimistic user engagement over time.

Figure \ref{tw_examples} presents some examples of tweets classified as positive and negative.



\subsection{Emotion Detection Results}

\begin{figure}[h!]
    \begin{subfigure}{.9\linewidth}
        \includegraphics[width=\textwidth]{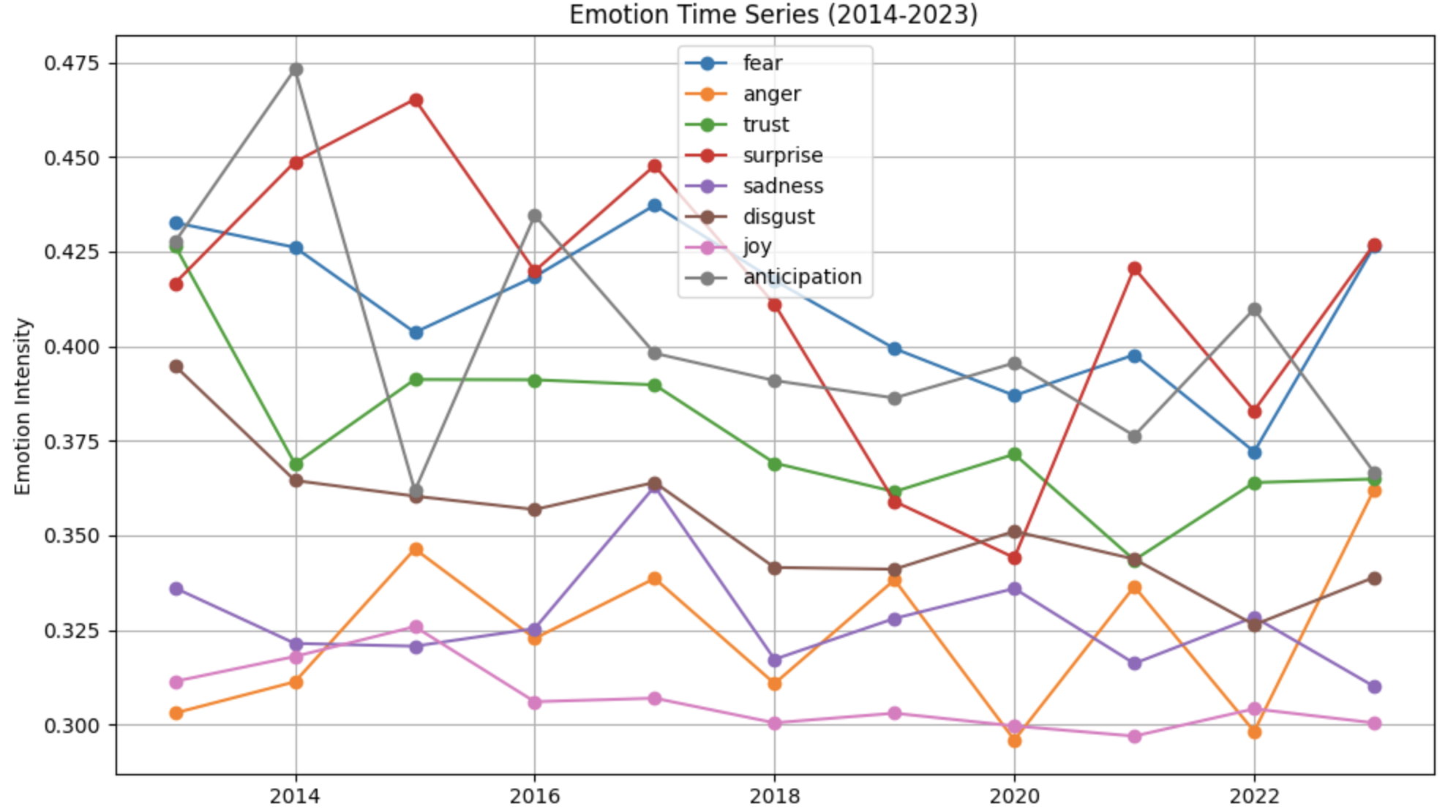}
        \caption{Twitter}
    \end{subfigure}%
    \hfill
    \begin{subfigure}{.9\linewidth}
        \includegraphics[width=\textwidth]{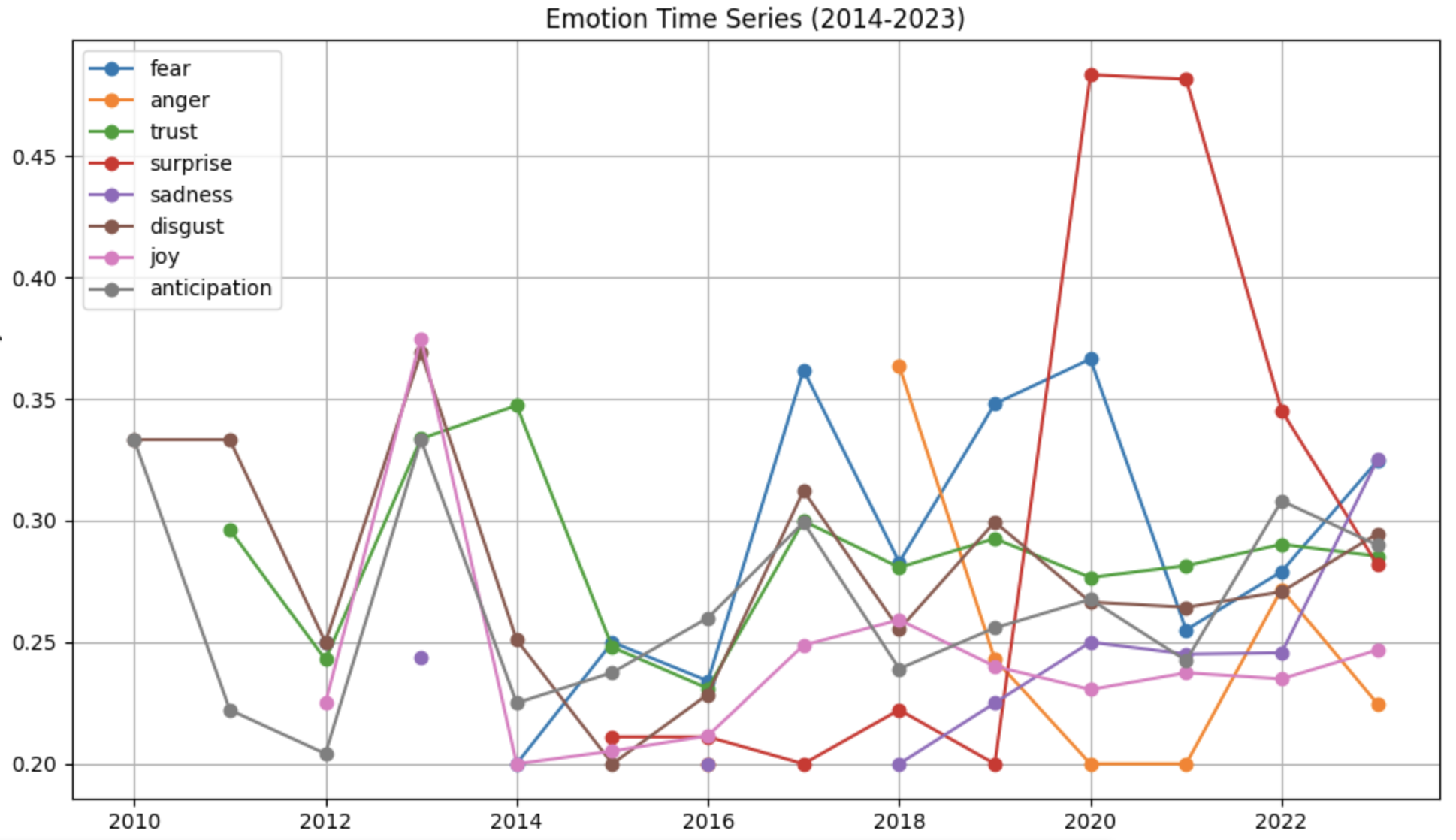}
        \caption{Reddit}
    \end{subfigure}%
    \hfill
    \begin{subfigure}{.9\linewidth}
        \includegraphics[width=\textwidth]{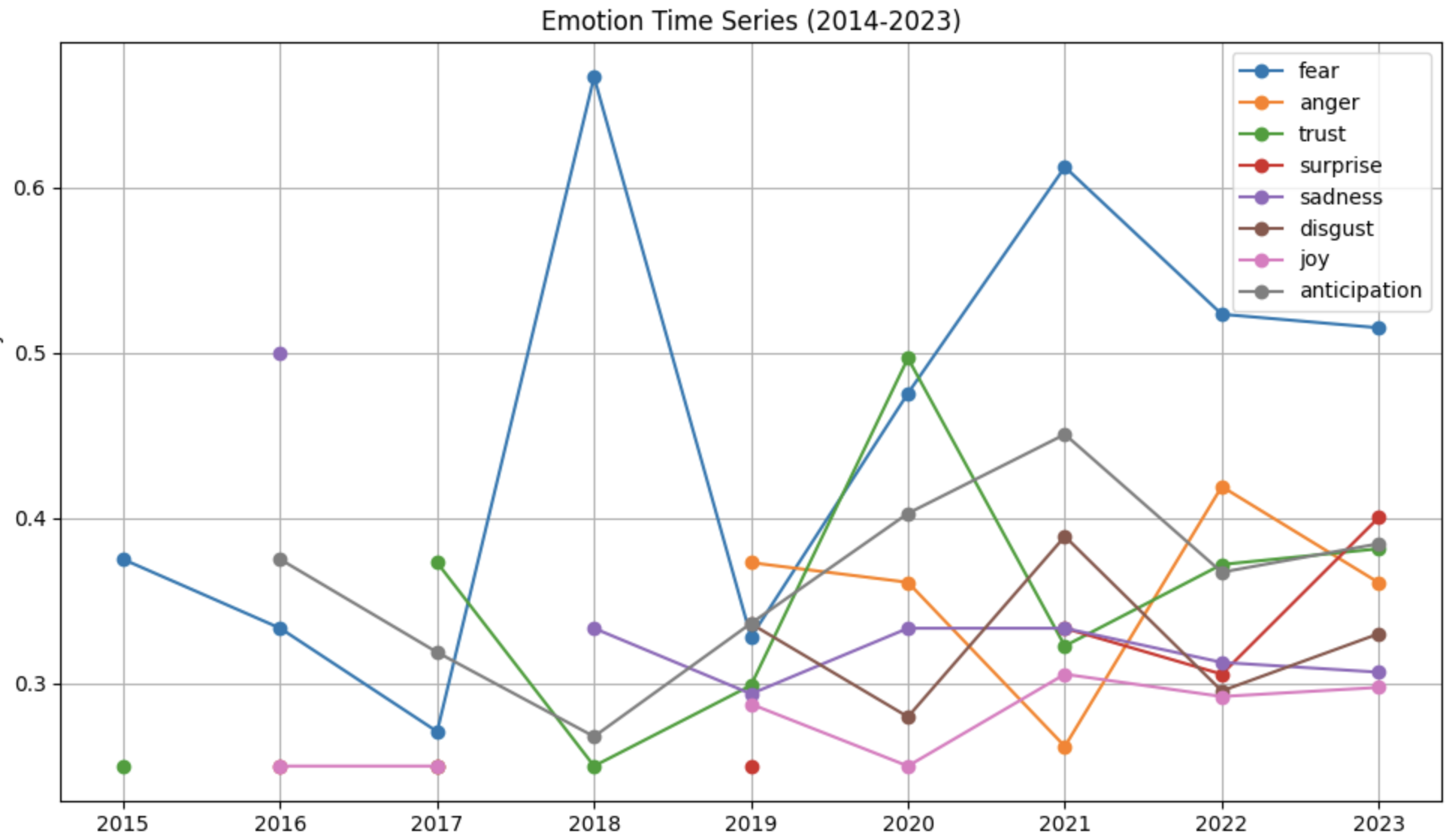}
        \caption{YouTube}
    \end{subfigure}%

    \caption{Mean emotion intensity scores over the years. For some emotions, the line is breaking due to the mean value of that emotion being lower than 0.25 in that year.}
    \label{fig:emotion_scores}
\end{figure}

Once we get only comments with negative sentiment scores, we input them into the NRClex tool to get emotion distribution in each comment. Comments with scores less than -0.1 were considered negative. Figure 
\ref{fig:emotion_scores} shows each social media platform's mean emotion intensity score over the years. We can clearly see that $\textit{fear}$, $\textit{trust}$, and $\textit{anticipation}$ were the most prevailing emotions over the years throughout all social media platforms. 

In the case of Twitter, all three emotions were growing simultaneously over time. $\textit{Trust}$ and $\textit{anticipation}$ peaked in 2020, which could be attributed to COVID-19 and tons of information being spread through social media during that period. Many users tended to agree and listen to posts from medical professionals, thus increasing trust and anticipation. Emotion $\textit{fear}$ also grows steadily with fewer fluctuations.

Unfortunately, YouTube has a major lack of information between 2014 and 2019. This happened because news channel accounts on YouTube started actively publishing videos related solely to the environment and climate change only recently, thus creating an information gap. However, we can observe an increasing number of $\textit{fear}$ emotions with its peak in 2022. Such a trend indicates growing user anxiety towards environmental challenges and acknowledging existing problems. $\textit{Trust}$, $\textit{anticipation}$ and $\textit{sadness}$ show a visible growth in the last three years. Both $\textit{trust}$ and $\textit{anticipation}$ could be attributed to the nature of the data, as it was scraped from the YouTube accounts of popular news channels, indicating users' trust towards information released on the official news channel accounts.



\subsection{Topic Modeling}

\begin{table*}[!htbp]
 \caption{Topic clusters per social media platform and prevailing emotion.}
  \setlength{\arrayrulewidth}{0.5mm}
    \setlength{\tabcolsep}{3pt}
    \renewcommand{\arraystretch}{1.5}
  \begin{tabular*}{\textwidth}{@{\extracolsep{\fill}}lcccc}
    \toprule
     & \textbf{Overall} & \textbf{Fear} & \textbf{Trust} & \textbf{Anticipation} \\
    \toprule
    Twitter & 
        \begin{tabular}{c}
         Climate change \\ Wastewater  \\  Plastic recycling\\ Biodiversity ecology\\ $CO_2$ and oxygen \\
        \end{tabular}  & 
        \begin{tabular}{c}
         Climate change \\Biodiversity \\ Plastic recycling  \\  Air quality\\ Foodwaste and vegan \\
        \end{tabular} & 
        \begin{tabular}{c}
         Climate and energy \\Biodiversity and food \\ Green energy  \\  $CO_2$ and air\\ \\
        \end{tabular} & 
        \begin{tabular}{c}
         Emissions \\Energy and carbon \\ Ecology conservation  \\  Eco-friendly\\ Air quality \\
        \end{tabular} \\
     \midrule
    Reddit & 
        \begin{tabular}{c}
         Plastic \\ Air quality level  \\
        \end{tabular}  & 
        \begin{tabular}{c}
         Air and cars \\ Single plastic use \\ Climate change \\
        \end{tabular} & 
        \begin{tabular}{c}
         Plastic recycling \\ Air quality monitor \\
        \end{tabular} & 
        \begin{tabular}{c}
         Store use plastic\\ Eco friendly\\
        \end{tabular} \\
    \midrule
    YouTube & 
        \begin{tabular}{c}
         Climate and people \\ Plastic bags  \\  Private jets\\ 
        \end{tabular}  & 
        \begin{tabular}{c}
         Climate and people \\Private jets \\  Hot summer  \\
        \end{tabular} & 
        \begin{tabular}{c}
         Carbon emissions \\ Change and warming \\ Energy and emissions  \\ 
        \end{tabular} & 
        \begin{tabular}{c}
         Oil and plastic \\Climate change and ice \\ Time to start\\  
        \end{tabular} \\
    \hline
  \end{tabular*}
  \label{tab:topics}
\end{table*}

As mentioned, Topic Modeling was performed by BERTopic \cite{grootendorst2022bertopic}. To get an insight into all scraped comments, we fed the cleaned sentences into BERTopic. The results correspond to the "Overall" column in Table \ref{tab:topics}. The comments from each platform were clustered into a maximum of five topics. The topics reflect the most common themes for the discussion on each social media platform. Clearly, \textit{Climate change} represents the biggest topic cluster across all platforms. \textit{Air quality}, \textit{Emissions}, \textit{Plastic}, and \textit{Recycling} also seem to be common discussions in all three datasets. 

Within \textit{Climate change} topic, users raise their concerns about global warming and criticize governmental ignorance. It was also common for users to demonstrate skepticism towards climate change's real threat; some users justified it as the natural process for the planet and named the opposite opinion a "Climate hysteria."

\textit{Air quality} and \textit{Emissions} seems to be mostly addressing vehicles and transport. Many users discussed the emissions released into the atmosphere during flights. On YouTube, users mostly encouraged to switch from gas and petrol-fueled cars to electric and hybrid alternatives.

\textit{Recycling} topic includes discussions about the ban on single-use plastic and companies shifting to recycled plastic or paper. Some users raised their concerns about whether recycling plastic individually will have any effect or whether the big companies should take responsibility and incorporate green practices.

On YouTube, several users expressed the irony of world leaders flying on private jets to the summit to discuss climate change. Such discussions were likely sparked by news reports about outcomes of the "Climate Summit" or similar events. 

Within Twitter, discussions about biodiversity were also popular. Users worry about the receding planet's biodiversity and the role of human activity in that problem.

Overall, we could see similar topics and patterns across all three social media platforms, indicating that the platform does not have a notable effect on user comments and discussions. It was also observed that there is still a good portion of users who consider climate change propaganda and its effects as a natural process for the planet that has been occurring before. Table \ref{table:comment_topic} provides comment examples per social media platform and the associated topic and emotion.

\begin{table*}[!htbp]
\caption{Sample comments with their corresponding topic, emotion and the social media platform}
\setlength{\arrayrulewidth}{0.5mm}
\setlength{\tabcolsep}{3pt}
\renewcommand{\arraystretch}{1.5}
\begin{tabularx}{\textwidth}{ Xccc } 
\hline
\textbf{Comment} & \textbf{Emotion} & \textbf{Topic}  & \textbf{Platform}\\
\hline
Investigation done by @CTVToronto reveals poor air quality in \#TDSB classrooms could contribute to headaches and sleepiness  & Anticipation & Air quality and emissions & Twitter\\
A Series of NASA Satellite Images Showcasing the Improvement in United States Air Quality From 2005 to 2011  & Trust & Air quality and emissions & Twitter\\
Some were concerned the city’s air quality, made worse by commercial expansion, would worsen allergies and asthma  & Fear & Air quality and emissions & Twitter\\
So I got an air quality monitor showing PM2.5 levels of 150 \textmu g/m3. That can't be good right? My cat has asthma attacks constantly and I'm wondering if its linked  & Trust & Air quality and emissions & Reddit\\
Smoking flares do not count when reducing emissions & Trust & Air quality and emissions & YouTube\\

How in the fuck can people look at the world around us and think climate change isn't real & Trust & Climate change & Twitter\\
When they ban jet flights to the Maldives and make everyone sail or swim there we'll their Global Warming hysteria seriously & Fear & Climate change & Twitter\\
Duh climate has always changed since 1990’s it’s nothing new it goes from cold , warm , dry , tornadoes, thunderstorms, rain , earthquakes & Trust & Climate change & YouTube\\
It seems so terrible. Increasingly the real harm of climate change is getting bigger, year by year & Fear & Climate change & YouTube\\
Ironic that everybody is worrying about climate change now when we knew decades ago this was coming & Anticipation & Climate change & YouTube\\
The newest climate report from the Intergovernmental Panel on Climate Change (IPCC) says that climate change is "unequivocally" caused by humans and warns that global temperatures are expected to reach a significant warming milestone in the next 20 years & Fear & Climate change & Reddit\\

The Climate Summit burned a hole in the Ozone with all the private jets that flew there & Fear & Private jets & YouTube\\

As much as 40\% of the food produced in America ends up at the dump, off-gassing methane and contributing to climate change & Fear & Foodwaste & Twitter\\
Ppl are more comfortable w/ Recycling b/c Reducing and Reusing challenge capitalism too much  & Anticipation & Recycling and eco-friendly & Twitter\\
Sustainability and environment expert, Dr. Jody Tishmack, provides an insightful explanation of our current water crisis, the impact it will have on society, and the solutions needed to manage it & Trust & Recycling and eco-friendly & Reddit\\

\hline
\end{tabularx}
\label{table:comment_topic}
\end{table*}

We examined data from three social networks and observed common trends with slight differences, likely stemming from the distinct nature of each platform, different user demographics, communication styles, etc. For instance, YouTube's inclusion of visual data in the form of videos may impact the associated comments.

\subsection{Positivity bias test}

\begin{figure*}[!htbp]
    \begin{subfigure}{0.33\textwidth}
        \includegraphics[width=\textwidth]{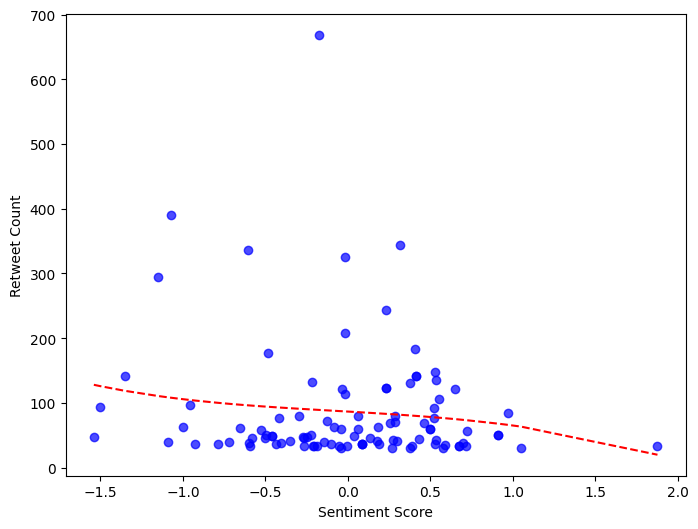}
        \caption{Twitter}
    \end{subfigure}%
    \hfill
    \begin{subfigure}{0.33\textwidth}
        \includegraphics[width=\textwidth]{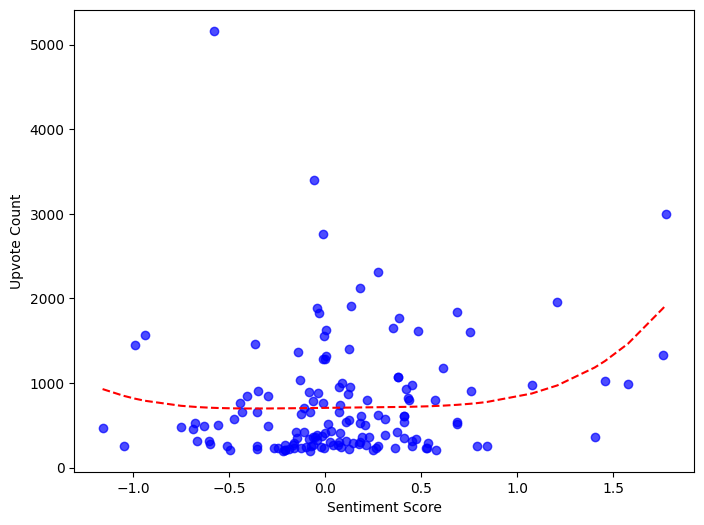}
        \caption{Reddit}
    \end{subfigure}%
    \hfill
    \begin{subfigure}{0.33\textwidth}
        \includegraphics[width=\textwidth]{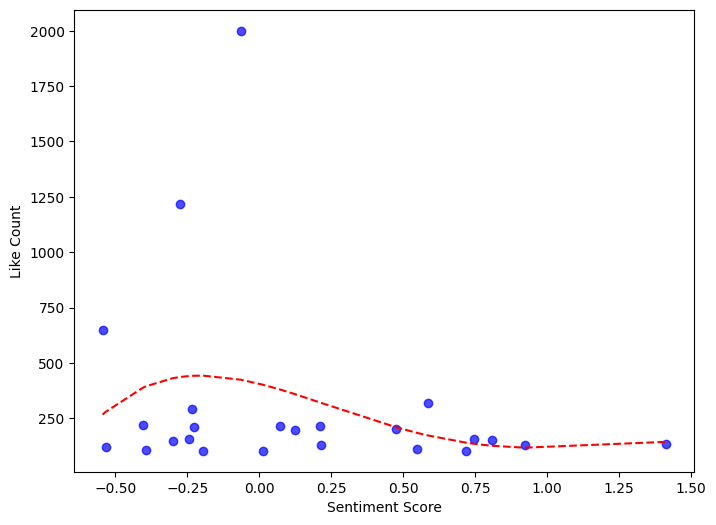}
        \caption{YouTube}
    \end{subfigure}%

    \caption{The number of retweets/upvotes/likes received by extremely popular posts as a function of the sentiment score represented in them. We can see the positivity bias on Reddit and the negativity bias on Twitter and YouTube.}
    \label{biases}
\end{figure*}
\begin{figure}[h]
  \centering
  \includegraphics[width=0.5\textwidth]{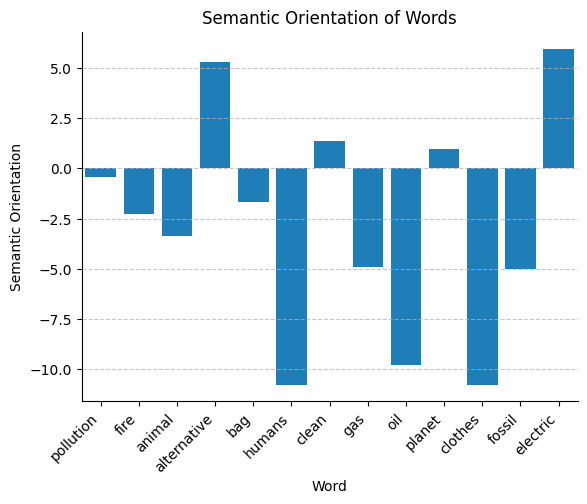}
  \caption{Words semantic orientation scores.}
  \label{fig:semantic_orientation}
\end{figure}
Whether more positive or more negative posts are more popular on each platform? The impact of sentiment on the virality of content in social media has been a subject of considerable interest. While some studies indicate a tendency for negative content to be shared more frequently \cite{negbias, negbias2}, contrasting findings suggest that positive information on social media is often more likely to garner likes and retweets \cite{diffusion1}. Using our collected dataset, we aim to investigate and analyze the relationship between sentiment and content sharing.

For the analysis, we filtered out only viral, extremely popular tweets, Reddit posts, and YouTube comments, with at least 30 retweets (RT $\ge$ 30) for Twitter, at least 200 upvotes for Reddit, and 100 likes for YouTube comments. As each social media platform is unique, the filtering parameters also differ. Figure \ref{biases} shows viral media posts' retweets/upvotes/likes as a function of their sentiment score. The data was fitted into a polynomial function. The graph demonstrates the positivity bias on Reddit and the negativity bias on Twitter and YouTube.

Our findings partly back up previous research on the influence of sentiment on information spread \cite{diffusion1}. Based on their research, positive messages are more likely to be shared and liked due to a phenomenon known as positivity bias. In our case, only Reddit confirms the positivity bias, probably because the context of the environment is mostly negative.

Next, the average number of retweets for negative tweets is larger (7.37 and 5.41 for negative and positive tweets, respectively) for Twitter. At the same time, the average number of likes/upvotes for negative comments is smaller for YouTube (23.45 and 30.95 for negative and positive comments, respectively) and Reddit (56.11 and 60.22 for negative and positive posts/comments, respectively). 

\subsection{Context-specific semantic orientation of words}

The analysis of semantic orientation (SO) scores for some context-specific keywords yielded interesting insights into the sentiment associations of these words. The SO scores indicate the sentiment associations within the context of the sentiment analysis. Based on Figure \ref{fig:semantic_orientation}, among these keywords, \textit{"alternative"} and \textit{"electric"} stand out with notably positive scores, suggesting strong positive associations, potentially indicative of favorable perceptions towards alternative solutions and electric-related aspects. Conversely, words like \textit{"fire", "animal", "humans", "oil", "clothes"} and \textit{"fossil"} exhibit extremely negative sentiment scores, hinting at severe negative associations within the context, perhaps highlighting concerns about environmental degradation, social issues, or adverse impacts. 

Keywords such as \textit{"pollution", "bag", "gas"} demonstrate a small degree of negativity, indicating concerns or associations that lean towards negative aspects within the context.

\textit{"Clean"}and \textit{"planet"} show moderately positive scores, hinting at positive associations, potentially linked to cleanliness or considerations for the well-being of the planet. 

These semantic scores collectively reflect a nuanced landscape of sentiments and associations surrounding these keywords within the specified domain, showcasing a wide range from deeply negative to strongly positive perceptions and concerns.

Analyzing the sentiment analysis scores of specific keywords in the dataset helps us gain a better understanding of how they are perceived and associated with sentiment. This provides valuable insights into subtle sentiment patterns and serves as a foundation for further discussions and interpretations in the field of sentiment analysis in environmental contexts.





\subsection{Accuracy Evaluation}

To assess the effectiveness of our method, we use the human-annotated dataset previously discussed, and the well-known sentiment analysis models VADER \cite{vader} and spaCy \cite{spacy}, and Senti \cite{senti, senti2}.

We examine the classification algorithms using metrics like $F_1$ score, Accuracy, Precision, and Recall. 

Precision represents the ratio of the number of true positive predictions to the total number of positive predictions.
\newline
\[Precision = \frac{TP}{TP + FP}\]

Recall shows accurate positive predictions when compared to the total number of actual positives\cite{metrics}: 
\newline
\[Recall = \frac{TP}{TP + FN}\]

$F_1$ score is \cite{metrics}:\newline
\[F_1 = 2 * \frac{Precision * Recall}{Precision + Recall} \]

Accuracy is a measure that represents the ratio of correct predictions to the total number of predictions made \cite{metrics}:
\newline
\[Accuracy =  \frac{TP +TN}{TP +TN+FP+FN} \]


\begin{table*}[!htbp]
\caption{Sentiment prediction accuracy, precision, recall, F1 for the proposed PMI-based, VADER, spaCy, senti classifiers. We also provided the evaluation for annotator $EA_1$, who has a Ph.D. in ecology, just for comparison.}
\setlength{\arrayrulewidth}{0.5mm}
\setlength{\tabcolsep}{35pt}
\renewcommand{\arraystretch}{1.5}
\begin{tabularx}{\textwidth}{ lcccc } 
\hline
 & \textbf{Accuracy} & \textbf{Precision}  & \textbf{Recall}  & \textbf{F1}\\
\hline
Proposed PMI classifier & 0.65 & 0.65 & 0.65 & 0.64 \\
VADER classifier & 0.64 & 0.80 & 0.64 & 0.71  \\
spaCy classifier & 0.44 & 0.44 & 1.00 & 0.61 \\
Senti classifier & 0.57 & 0.72 & 0.57 & 0.64 \\
$EA_1$, PhD in Ecology & 0.90 & 0.94 & 0.9 & 0.92   \\

\hline
\end{tabularx}
\label{table:accuracy}
\end{table*}

\begin{table*}[!htbp]
\caption{Results of sentiment analysis for 100 tweets: VADER, spaCy, Senti, Expert Annotator, and Pmi-based methods.}
\setlength{\arrayrulewidth}{0.5mm}
\setlength{\tabcolsep}{3pt}
\renewcommand{\arraystretch}{1.5}
\begin{tabularx}{\textwidth}{ Xcccccccc } 
\hline
\textbf{ Tweet} & \textbf{VADER} & \textbf{spaCy}  & \textbf{Senti}  & \textbf{EA1} & \textbf{PMI score} &\textbf{PMI sent.}&\textbf{Annotated AVG}&\textbf{Annotated label}\\
\hline
Imagine 2050, when there's  absolutely no denying terrible  reality of climate change, and  it's too late, WHY DIDN'T THEY ACT 50 YRS EARLIER? 
& Positive  & Positive  & Negative & -5 (Negative) & -0.19 & Negative& -4.63 & Negative\\
Imagine Mexico City with all electric cars and no chocking smog 
& Negative  & Positive  & Positive &3 (Positive) & -2.31 & Negative & 3.25 & Positive \\
Lets take action people... Save energy, cut down red meat and practice zero waste
& Positive  & Positive  & Neutral &5 (Positive) & 1.7 & Positive & 3.88 & Positive \\
Make a difference! Join the fight to defend the teaching of evolution and climate change
& Negative  & Positive  & Negative &4 (Positive) & -0.33 & Negative & 2.75 & Positive \\
I want to go zero waste but there's no such thing as zero waste art supplies, everything comes in aluminum tubes and I need to use lots of paper towels and ...aren't good for the planet either:( 
& Negative  & Positive  & Positive &-5 (Negative) & -0.19 & Negative & -2.75 & Negative \\

\hline
\end{tabularx}
\label{table:compare}
\end{table*}

Table~\ref{table:accuracy} shows evaluation results for each method.  Sentiment prediction accuracy, precision, recall, and F1 scores were calculated for PMI-based, VADER, spaCy, and senti classifiers. Additionally, we provided evaluation results for annotator $EA_1$, who holds a Ph.D. in ecology and demonstrated the lowest outliers rate among all annotators (12\%) for comparison.
Based on the results presented in Table~\ref{table:accuracy}, it can be observed that our classifier (0.65) outperforms VADER (0.64), Senti (0.57), and spaCy (0.44) for the selected context. However, it still falls short of the accuracy achieved by individual human raters with specialized context knowledge (0.90). Although VADER is known to be sensitive to social media lexicon, its accuracy decreases when used in domain-specific environmental tweets. The dataset used for training has limited variability and context-specific vocabulary, which is why a PMI-based classifier performs better.

Table \ref{table:compare} compares the Mutual-Information scoring function to the VADER, spaCy, Senti and expert annotator scores, along with the example tweets from the labeled dataset.

The last row in Table \ref{table:compare} represents the most controversial tweet with the highest standard deviation(std = 4.22) in our manual annotation experiment. Despite "zero waste" typically being associated with positivity, the tweet explores the challenges of adhering to this concept in the modern world. The heightened standard deviation reflects the varying interpretations among annotators, underscoring the nuanced sentiment and real-world complexities embedded in the discussion of "zero waste."


\section{Discussion}







Let's explore how our research findings align with previous studies. Our findings diverge notably from recent research results \cite{ROSENBERG2023101287}, particularly in the sentiment analysis within Twitter data. Recent studies have underscored a prevailing tendency towards tweets with positive sentiments. For instance, their study showcased that 19.7\% of tweets carried negative sentiments for the year 2021. However, our analysis presents a contrast, revealing a substantially higher prevalence of negative sentiments, accounting for 54.7\% of tweets within the same year.

\cite{Effrosynidis2022} conducted a study involving clustering Twitter data into various topics, aligning with our own topic modeling efforts. Notably, themes such as \textit{"climate change"} and \textit{"carbon emissions"} surfaced in both studies, underscoring their mutual support.

Conversely, an analysis by \cite{article} on emotion detection in tweets produced divergent outcomes from our findings. While their study highlighted anger as the predominant emotion, our research identifies anticipation and fear as the dominant emotions observed in the analyzed data.

It's worth highlighting that similar studies like \cite{ROSENBERG2023101287} have consistently revealed a pattern of relatively low accuracy in sentiment classification for environmental tweets. For instance, the model employed by this study (VADER) achieved 56\% of accuracy, whereas our classifier achieved a higher accuracy of 65\%. This underscores the ongoing necessity for enhanced support in refining sentiment classification models to identify sentiments within environmental data effectively.


\section{Conclusion}


This study demonstrated the value of sentiment analysis in understanding the public perception of environmental topics over a decade. We explored multiple social networks for a broader perspective, included emotion detection to capture a wide range of emotional responses, and employed topic modeling techniques to identify specific environmental topics.

Our findings show that negative environmental tweets are much more common than positive or neutral ones (X\% vs Y\% and Z\% on average). Climate change is the primary topic across all social media platforms, emphasizing its widespread global concern. Additionally, discussions on \textit{air quality, emissions, plastic, }and \textit{recycling} consistently appear in all datasets, highlighting their ongoing relevance. The prevailing emotions expressed in environmental tweets are \textit{fear, trust} and \textit{anticipation}, indicating public reactions' diverse and complex nature. 


These findings contribute to understanding how social media platforms perceive and discuss environmental topics. The results can be used to inform policymakers, organizations, and governments about the public's priorities, concerns, and suggestions related to environmental issues. This can lead to the formulation of more effective environmental policies.


 Our study faced a key limitation: low sentiment analysis accuracy in environmental tweets. This challenge arises from the predominantly negative tone surrounding climate change discussions, making it challenging to identify nuanced positive sentiments. Additionally, we have observed that such tweets often contain irony and sarcasm, which can also hinder accuracy.




As for future works, we plan to incorporate multimodal analysis and use images besides text. We also aim to compare and contrast social media sentiments with sentiments expressed in traditional media, like news articles, TV, and radio.


\section*{Acknowledgements}
We extend our appreciation to Dr. Aiymgul Kerimray, PhD in ecology, and our team of annotators for their dedicated efforts in annotating the environmental tweets.

\bibliography{export}

\end{document}